%% file: main.tex
\documentclass[10pt,twocolumn,letterpaper]{article}

\usepackage[pagenumbers]{cvpr} 

\input{preamble}

\usepackage{pifont}
\usepackage{tabularx}
\usepackage[export]{adjustbox}
\usepackage{multirow}
\usepackage{makecell}
\usepackage{xcolor,colortbl}
\usepackage{dsfont}
\usepackage{color} 
\definecolor{LightGray}{RGB}{220, 220, 240} 

\definecolor{cvprblue}{rgb}{0.21,0.49,0.74}
\usepackage[pagebackref,breaklinks,colorlinks,citecolor=cvprblue]{hyperref}

\def\paperID{*****} 
\def\confName{CVPR}
\def\confYear{2024}

\title{Low-Rank Few-Shot Adaptation of Vision-Language Models}

\author{Maxime Zanella\thanks{Corresponding author: maxime.zanella@uclouvain.be\\This work was partly supported by the Walloon Region
(Service Public de Wallonie Recherche, Belgium) under grant
n°2010235 (ARIAC by DigitalWallonia4.ai).}\\
UCLouvain \quad UMons 
\and Ismail Ben Ayed
\\ \'ETS Montr\'eal \and 
{\tt code: \url{https://github.com/MaxZanella/CLIP-LoRA}}
}

\begin{document}
\maketitle

\input{text/0_abstract}
\input{text/1_intro}
\input{text/2_related_work}
\input{text/3_method}
\input{text/3bis_lora}
\input{text/4_results}

\input{text/5_ablation}

\input{text/6_conclusion}
{
    \small
    \bibliographystyle{ieeenat_fullname}
    \bibliography{mybib}
}

\input{text/X_appendix}


\end{document}

%% file: preamble.tex
\usepackage[dvipsnames]{xcolor}

\usepackage{subcaption}

%% file: text/0_abstract.tex
\begin{abstract}
Recent progress in the few-shot adaptation of Vision-Language Models (VLMs) has further pushed their generalization capabilities, at the expense of just a few labeled samples 
within the target downstream task. However, this promising, already quite abundant few-shot literature has focused principally on prompt learning and, to a lesser extent, on adapters, overlooking the recent advances in Parameter-Efficient Fine-Tuning (PEFT). Furthermore, existing few-shot learning methods for VLMs often rely on heavy training procedures and/or carefully chosen, task-specific hyper-parameters, which might impede their applicability. In response, we introduce Low-Rank Adaptation (LoRA) in few-shot learning for VLMs, and show its potential on 11 datasets, in comparison to current state-of-the-art prompt- and adapter-based approaches. Surprisingly, our simple CLIP-LoRA method exhibits substantial improvements, while reducing the training times and keeping the same hyper-parameters in all the target tasks, i.e., across all the datasets and numbers of shots. Certainly, our surprising results do not dismiss the potential 
of prompt-learning and adapter-based research. However, we believe that our strong baseline could be used to evaluate progress in these emergent subjects in few-shot VLMs.
\end{abstract}

%% file: text/1_intro.tex
\section{Introduction}
Vision-Language Models (VLMs), such as CLIP~\cite{clip}, have emerged as powerful tools for learning cross-modal representations~\cite{clip, align, filip, lit, li2022grounded, vlm_review}. Pre-trained on extensive collections of image-text pairs with a contrastive objective~\cite{clip}, VLMs learn to align 
these two modalities, enabling zero-shot prediction by matching the visual embeddings of the images and text descriptions (prompts) representing the target tasks. 
This joint representation space of visual and textual features has also opened up new possibilities in the \textit{Pre-training, Fine-tuning, Prediction} paradigm, through adaptation with very limited amounts of task-specific, labeled data~\cite{coop, prograd, tip-adapter}, i.e., {\em few-shot adaptation}. Nonetheless, their efficacy often relies on the use of transformer-based architectures~\cite{vaswani2017attention}, where larger variants significantly outperform smaller ones. For instance, in CLIP, the ViT-L/14 model significantly surpasses the ViT-B/16 version by over 6\% in accuracy on ImageNet~\cite{imagenet}---a disparity that persists even following few-shot adaptation of the models. This underscores the need for efficient vision-language fine-tuning methods that are scalable to large models. The recent and emergent few-shot vision-language literature, although quite abundant already, has so far overlooked the computational overhead and memory footprint of fine-tuning strategies. This is the case of both the so-called {\em adapters}, which equip the models with additional trainable parameters~\cite{tip-adapter}, and popular {\em prompt-learning} methods~\cite{coop}, which fine-tune the input text prompts. This can increase the computational demand and the size of these already substantial models.\\
\par These questions surrounding the fine-tuning stage have triggered wide interest in NLP, where the increase in the sizes of foundational models is going at a fast pace, with some models boasting over 176 billion parameters~\cite{workshop2022bloom}, and even up to 540 billion~\cite{chowdhery2023palm}. To address these challenges, {\em Parameter Efficient Fine-Tuning (PEFT)} methods, which attempt to fine-tune only small amounts of parameters (in comparison to the original large models), have gained substantial attention~\cite{lialin2023scaling}. 
Popular PEFT methods include selecting a subset of the existing parameters~\cite{zaken2022bitfit}, adding small trainable modules called adapters~\cite{rebuffi2017learning, houlsby2019parameter, NEURIPS2021_081be9fd, chen2022adaptformer,lian2022scaling}, or adding trainable (prompt) tokens~\cite{prompt_tuning, li2021prefix, jia2022visual}. Recently, a novel approach consisting of solely fine-tuning low-rank matrices called Low-Rank Adaptation (LoRA) has appeared as a promising and practical method~\cite{hu2021lora}. PEFT approaches, such as LoRA, have democratized the fine-tuning of large-language models, enabling even the management of billion-parameter models on a single GPU~\cite{dettmers2024qlora}. Far from merely enhancing computational efficiency, empirical evidence has shown that, in the large-scale fine-tuning setting, LoRA could match or even exceed the performance of updating all model's parameters~\cite{hu2021lora}. 
{\em Although very promising/popular, and quite surprisingly, this fast-growing PEFT literature~\cite{hu2021lora, valipour2022dylora, zhang2022adaptive, kim2023hydra, zi2023delta, dettmers2024qlora} has found little echo in few-shot vision-language, where the dominant approaches have mainly focused on prompt tuning~\cite{coop, cocoop, kgcoop, lasp, prograd, plot} or adapters~\cite{tip-adapter, clip-adapter, yu2023task}}.\\
\par The original CLIP paper demonstrated that better textual descriptions could greatly impact the zero-shot prediction~\cite{clip}. This observation has been a strong motivation for the 
emergence of prompt tuning~\cite{prompt_tuning}, a strategy that has been widely adopted within the vision-language community, following the seminal work of CoOp~\cite{coop}. Indeed, CoOp popularized prompt tuning in the setting of few-shot VLMs. This has triggered a quite substantial recent literature focusing on improving prompt-learning performances for VLMs, in both
the few-shot~\cite{coop, cocoop, proda, variational, kgcoop, lasp, prograd, plot} and unsupervised settings~\cite{upl, tpt, difftpt, zanella2024testtime}. While prompt learning methods improve the zero-shot performances, they incur heavy computational load for fine-tuning and might be hard to optimize, since every gradient update of the input requires back-propagating through the entire model; see the training times in Table \ref{tab:runtime}. 

Alongside this expanding prompt-tuning literature, there
has been a few attempts to propose alternative approaches for few-shot VLMs, generally relying on adapters~\cite{clip-adapter, tip-adapter, yu2023task}. However, the performances of such adapters depend strongly on a set of hyper-parameters (such as the learning rate, number of epochs, or parameters controlling the blending of image and text embeddings)~\cite{Julio-ArXiv23}, which have to be found specifically for each target dataset. This is done via intensive searches over validation sets, requiring additional labeled samples and incurring computational overhead, which reduces their portability to new tasks. 

\paragraph{Contributions.} In this work, we investigate the deployment of Low-Rank Adaptation (LoRA) in the context of few-shot VLMs, an emergent, already quite abundant literature 
dominated by prompt-learning and adapter-based strategies. We thoroughly examine different design choices for deploying LoRA in this context, namely, the choices of the encoders (vision, language or both), of the specific weight matrices to adapt, and of the rank of the matrices. We conduct comprehensive empirical ablations and comparisons, over 11 datasets, emphasizing the best design choices for our baseline and juxtaposing it to the existing state-of-the-art prompt- and adapter-based methods. Surprisingly, our LoRA baseline beats the state-of-the-art in few-shot VLMs by important margins, while reducing the computational overhead. Furthermore, it relaxes the need for intensive searches of the hyper-parameters over dataset-specific validation sets, maintaining a consistent hyper-parameter configuration across all the target tasks. While our surprising results do not invalidate the promise of prompt-learning and adapter-based strategies, we believe this strong baseline could be used to evaluate progress in these emergent subjects in few-shot VLMs.





%% file: text/2_related_work.tex
\section{Related work}
\input{figures/peft}
\par \textbf{Parameter-Efficient Fine-Tuning (PEFT).} PEFT seeks to reduce the high expense of fine-tuning large-scale models by concentrating on (re-)training a relatively small number of parameters. These techniques can be categorized into four groups, primarily distinguished by the choice of parameters to train~\cite{lialin2023scaling}. This often results in a trade-off among memory footprint, computational overhead, and performance. A summary is depicted in Figure \ref{fig:peft}.
\par The most straightforward way to avoid full fine-tuning is through {\em selective} methods, which focus on a subset of the existing model weights. Among these, BitFit~\cite{zaken2022bitfit} fine-tunes only the biases of both the attention and MLP layers in the transformer blocks, while other approaches prune the model to create a task-specific subset of parameters~\cite{guo2021parameter, holmes2021nxmtransformer}. 
\par Secondly, adapters integrate additional trainable modules into the original frozen architecture~\cite{rebuffi2017learning, houlsby2019parameter, NEURIPS2021_081be9fd, chen2022adaptformer,lian2022scaling}; for example, by shifting and scaling deep features~\cite{lian2022scaling}. They also demonstrate their versatility and effectiveness in various tasks implying vision and language~\cite{sung2022vl}. Nonetheless, the primary drawback of using adapters is the additional number of parameters after adaptation, which can lead to higher inference latency, even though some recent works aim at mitigating this issue~\cite{pfeiffer2021adapterfusion}. 
\par Thirdly, there is prompt tuning or token-based tuning~\cite{prompt_tuning, li2021prefix, jia2022visual}, which involves adding learnable tokens either to the input or at intermediate sequences. This strategy has been particularly popular in vision-language for few-shot and zero-shot learning, replacing hard-to-design template prompts with learnable soft ones~\cite{coop, tpt}. Initially applied to textual prompts, recent works have extended this technique to train visual tokens within transformer-based architectures~\cite{jia2022visual}. This research direction has begun to spark interest in the few-shot vision-language field~\cite{khattak2023maple}. 
\par Finally, Low-Rank Adaptation (LoRA)~\cite{hu2021lora} adds low-rank matrices to explicitly represent weight changes while keeping the original parameters frozen. These explicit changes can then be merged with original weights prior inference, inducing no additional inference latency in comparison to the vanilla model. LoRA operates on the hypothesis that updates required for the fine-tuning process exhibit a low “intrinsic rank”~\cite{li2018measuring, aghajanyan2021intrinsic}, a property we can directly control with the rank of each weight change matrix. In this respect, LoRA can be viewed as an adapter approach, yet it offers the advantages of selective methods by providing a direct aggregation of its module, eliminating the need for extra parameters at the inference stage. Several versions of the original LoRA have since appeared, some focusing on making the rank adaptive for each matrix~\cite{zhang2022adaptive, valipour2022dylora}, others pushing its performance~\cite{zi2023delta, kim2023hydra, chavan2023one} or reducing its memory footprint through quantization~\cite{dettmers2024qlora, rajabzadeh2024qdylora}.
\paragraph{Few-shot learning in Vision-Language.} Large scale VLMs have shown excellent results in several vision tasks~\cite{vlm_review}. This success has created interest in developing adaptation techniques that capitalize on their general knowledge~\cite{fine_tuning_clip}. Among these, prompt tuning~\cite{prompt_tuning} has emerged as the primary method for adapting VLMs with few labeled data~\cite{coop, cocoop, proda, variational, kgcoop, lasp, prograd, plot}. CoOp~\cite{coop} optimizes learnable common continuous tokens attached to the class names, described as a context optimization. CoCoOp~\cite{cocoop} trains a neural network to generate instance-conditioned tokens based on the image. Further efforts like ProGrad~\cite{prograd} and KgCoOp \cite{kgcoop}, among others \cite{lasp}, guide prompts towards predefined handcrafted ones, for example, thanks to gradients projection~\cite{prograd} with the idea of preserving initial knowledge during learning. 
\par Among prompt-learning works, PLOT~\cite{plot} is one of the first to adapt jointly the text and image modalities. They propose to align learned prompts with finer-grained visual features through an optimal transport formulation. Note that this cross-modal adaptation is also a key factor in our approach, as discussed in Section \ref{sec:ablation}. Following a similar trajectory, MaPLe~\cite{khattak2023maple} introduces intermediate learnable tokens within both the vision and text encoders, while making them interdependent at each level. They further demonstrate that adapting both modality branches allows for more flexibility in the downstream tasks.
\par Adapter-based methods offer an alternative strategy and are increasingly studied in vision-language~\cite{clip-adapter, tip-adapter, yu2023task}. CLIP-Adapter~\cite{clip-adapter} learns visual adapters to combine adapted and original features. A few other methods propose to leverage the knowledge of these models while only accessing their final embedding state. Examples include parameter-free plug-in attention for the zero-shot scenario~\cite{calip}, or Tip-Adapter(-F)~\cite{tip-adapter} using a cache model in few-shot learning. In a similar vein, TaskRes~\cite{yu2023task} keeps the original text weights frozen and introduces task residual tuning to learn task-specific adapters built on the initial knowledge. 

%% file: figures/peft.tex
\begin{figure*}[t]
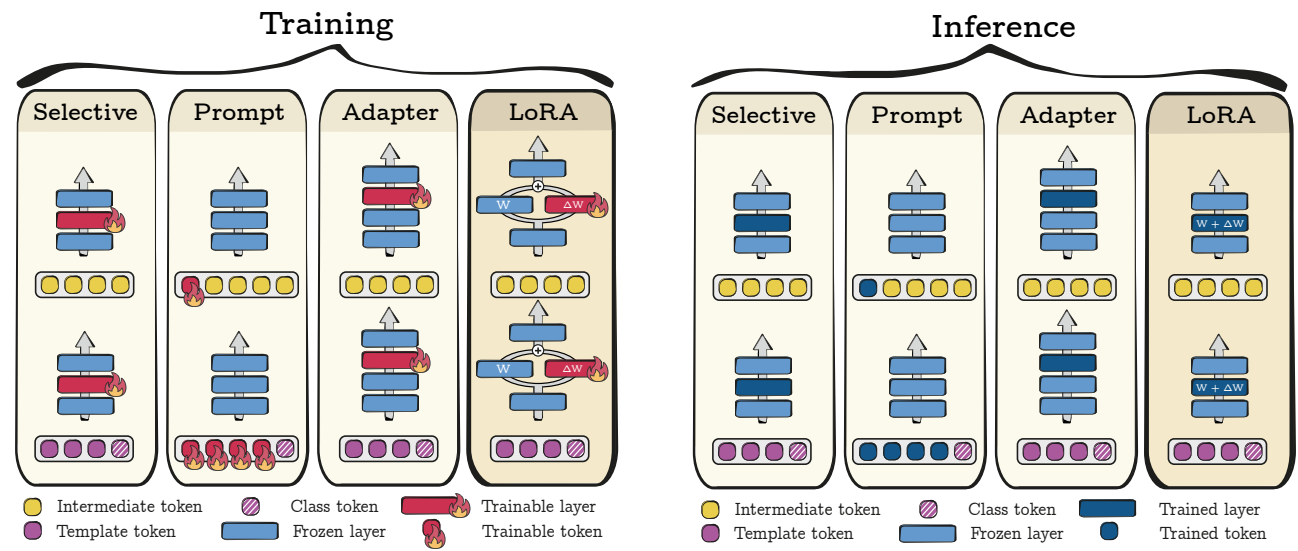

    \centering
    \begin{subfigure}[t]{0.47\textwidth}
    
        \centering
        \includegraphics[width=\linewidth]{figures/LoRAaseis.pdf}
        \caption{Prompt, Adapter and Low-rank techniques introduce extra parameters for training, which may potentially extend training duration and/or memory footprint in comparison to selective methods. However, they have the advantage of being more flexible and are often easier to use.}
    \end{subfigure}%
    ~ \hspace{0.5cm}
    \begin{subfigure}[t]{0.47\textwidth}
    
        \centering
        \includegraphics[width=\linewidth]{figures/LoRAbseis.pdf}
        \caption{Prefix and Adapter methods result in extra parameters after adaptation, potentially increasing inference time and memory footprint relative to the vanilla model. Conversely, LoRA merges newly trained low-rank matrices with the original frozen ones, eliminating additional parameters at inference.}
    \end{subfigure}
    \caption{Different categories of Parameter-Efficient Fine-Tuning (PEFT) methods during (a) training, and (b) inference.}
    \label{fig:peft}
\end{figure*}

%% file: text/3_method.tex
\section{Few-shot fine-tuning for VLMs}
This section provides a broad overview of recent few-shot fine-tuning methods designed for VLMs, summarized in Figure \ref{fig:peft}. First, let us introduce a 
few basic notations and definitions. 

When dealing with a classification task based on a vision-language model, and given a set of $K$ candidate classes, one creates textual descriptions, the so-called prompts~\cite{liu2023pre}, each corresponding to a class, e.g., ${\mathbf c}_k$ is the tokenized version of \texttt{a photo of a} [$\texttt{kth class name}$], $k=1,\dots, K$. Let ${\mathbf t}_k = \theta_t ({\mathbf c}_k)$ denotes the corresponding normalized (unit-hypersphere) textual embedding representation, with $\theta_t$ representing the parameters of the language encoder. Similarly, each image ${\mathbf x}_i$, $i =1, \dots, N$, is projected onto a normalized embedding space of the same dimension, using the visual encoder $\theta_v$:~${\mathbf f}_i = \theta_v ({\mathbf x}_i)$. 

The zero-shot prediction is the simplest form of adapting VLMs to a downstream task, which follows their pre-training procedure~\cite{clip}. Pairing each text embedding ${\mathbf t}_k$ with ${\mathbf f}_i$, the visual embedding of test 
image ${\mathbf x}_i$, one could measure their cosine similarity, yielding a prediction logit:
\begin{equation}
\label{logits-ik}
l_{i,k} = {\mathbf f}_i^\top {\mathbf t}_k. 
\end{equation}
This also yields a probabilistic prediction, in the form of a posterior softmax probability of class $k$ given test input ${\mathbf x}_i$:         
\begin{equation}
\label{zero-shot-prediction}
p_{i,k}= \frac{\exp (l_{i,k}/\tau)}{\sum_{j}^K \exp (l_{i,j}/\tau)}  
\end{equation}
where $\tau$ is a softmax-temperature parameter\footnote{Note that each CLIP version comes with a temperature scaling $\tau$, which is optimized along with the learnable parameters during pre-training.}. Hence, zero-shot classification of image ${\mathbf x}_i$ is done by finding the class with the highest posterior probability: $\hat{k} = \mathrm{argmax}_{k}~p_{i,k}$.
\input{tables/runtime}

In the few-shot setting, and to further adapt these models, we assume that we have access to $N/K$ labeled samples for each target class, the so-called {\em support} set. $N$ denotes the total number of support samples, and $N/K$ (the number of shots per class) is typically small (less than $16$). 
Let $y_{ik}$ denotes the one-hot encoded label for a labeled support image ${\mathbf x}_i$, i.e., $y_{ik}=1$ if $k$ is the class of image ${\mathbf x}_i$ and $0$ otherwise.
Then, we minimize the cross-entropy (CE) loss:
\begin{equation}
\label{CE-loss}
  - \frac{1}{N}\sum_{i=1}^{N}\sum_{k=1}^{K} y_{ik}\ln{p_{i,k}}
\end{equation}
This is done either (i) by fine-tuning the input prompts, ${\mathbf c}_k$, $k=1,\dots, K$, as in prompt-tuning methods following on from the pioneering work of CoOp~\cite{coop, cocoop, plot, kgcoop, khattak2023maple, prograd}; or (ii) by updating a set of additional parameters, as in adapters ~\cite{clip-adapter, tip-adapter, yu2023task}. 
Note that other methods propose to tune additional intermediate tokens~\cite{khattak2023maple}, which we include under the category `'prompt tuning" for a more general terminology. We will now detail the two current strategies used in VLMs: Prompt tuning (P) and Adapters (A).

\paragraph{Prompt tuning (P).} The way prompting is performed in VLMs could significantly impact the ensuing performances~\cite{clip}. 
To address this issue, soft prompt tuning~\cite{prompt_tuning, li2021prefix} optimizes text-input tokens, which could be extended to intermediate layers~\cite{li2021prefix}. Similarly, if a transformer-based~\cite{vaswani2017attention} architecture is used, these learnable tokens can be inserted in vision models~\cite{jia2022visual}. In the context of few-shot VLMs, the authors of~\cite{coop} introduced context optimization (CoOp), 
which constructs text input ${\mathbf c}_k$ as continuous trainable vectors:
\begin{equation}
\label{lesrneable-tokens}
{\mathbf c}_k=(\mathbf{v}_k^1, \dots, {\mathbf v}_k^M, [\mbox{class}_k])
\end{equation}
Where $M$ denotes a hyper-parameter, $( {\mathbf v}_k^l)_{1 \leq l \leq M}$ are trainable text tokens, and $[\mbox{class}_k]$ is a fixed token. The latter is the word embedding vector of the name of the $k^{th}$ class. These trainable vectors are updated as task-specific text prompts by using the 
standard supervised CE classification loss in Eq. \eqref{CE-loss}, along with the few-shot labeled samples. Prompt tuning has a clear advantage over adapter-based methods: They remove the need for heuristically choosing the text prompts~\cite{coop}, which are specifically engineered for each task, and whose choice might affect the performances significantly. 
While prompt-tuning methods improves significantly the performances of classification, they incur heavy computational load for fine-tuning and might be hard to optimize, since every gradient update of the text input requires back-propagating through the entire model (see the training times reported in Table~\ref{tab:runtime}).


\paragraph{Adapters (A).} 
Instead of updating the text prompts, another class of methods, called adapters, augment the pre-trained model with extra parameters while keeping the existing ones frozen \cite{houlsby2019parameter}. This provides an efficient way to control the number of trainable parameters. The idea has been recently explored in the few-shot vision-language setting \cite{clip-adapter,tip-adapter,yu2023task}. In this setting, adapters could be viewed as feature transformations, via some multi-layer modules, appended to the encoder's bottleneck. This enables to learn transformations blending image and text features, with logits taking the following form:
\begin{equation}
\label{logits-adapters}
l_{i,k}  = \theta_a ({\mathbf f}_i, {\mathbf t}_k)
\end{equation}
where $\theta_a$ denotes the additional trainable parameters of the adapter. These are fine-tuned by minimizing the CE loss in \eqref{CE-loss}, using the labeled support set but now with logits $l_{i,k}$ transformed by \eqref{logits-adapters}. 
For example, CLIP-Adapter \cite{clip-adapter} added a multi-layered perceptron to modify the features. Tip-Adapter \cite{tip-adapter} added a module, which 
evaluates class scores via some pairwise similarities between the features of the labeled samples, and integrates these scores with the embeddings of the text prompts. This class of methods  
reduce the computational load in  comparison with prompt-tuning techniques. However, as pointed out recently in the experiments in \cite{Julio-ArXiv23},
their performances seem to depend strongly on some key hyper-parameters that have to be adjusted specifically for each downstream task. This is is done via intensive searches over validation sets, requiring additional labeled samples \cite{tip-adapter} and incurring computational overhead, which reduces their portability to new tasks. 

%% file: tables/runtime.tex
\begin{table}[t]
\caption{Training time on 16-shots ImageNet task. Experiments were conducted on a single A100 80Gb with the original code provided by the authors. For PLOT++ the time reported includes the 2 training stages.}
\label{tab:runtime}
\centering
\resizebox{0.5\linewidth}{!}{
\begin{tabular}{lc}
\toprule
Method & Training time 
\\ \midrule 
CoOp (16) & 2h  \\
PLOT++ & 15h30   \\ 
ProGrad & 3h20  \\
\rowcolor{LightGray} CLIP-LoRA & 50 min.  \\
\bottomrule 
\end{tabular}}

\end{table}

%% file: text/3bis_lora.tex
\section{CLIP-LoRA}
Low-Rank Adaptation (LoRA)~\cite{hu2021lora} models the incremental update of the pre-trained weights as the product of two small matrices, ${\mathbf A}$ and ${\mathbf B}$, based on the idea of `'intrinsic rank'' of a downstream task. 
For an input ${\mathbf x}$, a hidden state ${\mathbf h}$, and a weight matrix ${\mathbf W} \in \mathrm{R}^{d1\times d2}$, the modified forward pass, following the application of a LoRA module, is:
\begin{equation}
    h = {\mathbf W} {\mathbf x} + \gamma \Delta {\mathbf W} {\mathbf x} = {\mathbf W} {\mathbf x} + \gamma {\mathbf B} {\mathbf A} {\mathbf x}
\end{equation}
where  ${\mathbf A} \in \mathrm{R}^{r\times d2}$, ${\mathbf B} \in \mathrm{R}^{d1 \times r}$, $\Delta {\mathbf W} \in \mathrm{R}^{d1\times d2}$ of rank r, with r typically $\ll \{d1, d2\}$, and $\gamma$ a scaling factor. Values in ${\mathbf A}$ are randomly initialized via Kaiming initialization while ${\mathbf B}$ is filled with zeros. This implies that there is no incremental update before training, and therefore, the output remains unchanged.\\
In the original LoRA paper, the low-rank matrices are applied on the attention matrices of transformer-based architectures~\cite{vaswani2017attention}. They typically consist of L stacked blocks, each containing a multi-head attention (MHA) module:
\[
    \text{head}_i=\text{Softmax}\left(\frac{{\mathbf x}{\mathbf W}_{q_i} ({\mathbf x} {\mathbf W}_{k_i})^T}{\sqrt{d}} \right)({\mathbf x}{\mathbf W}_{v_i})
\]
\[
    \text{MHA}({\mathbf x}) = \text{concat}(\text{head}_1,...,\text{head}_H){\mathbf W}_o
\]
where $d$ is a scaling factor and ${\mathbf W}_{K_i}$, ${\mathbf W}_{Q_i}$, ${\mathbf W}_{V_i}$, ${\mathbf W}_o$ are weight matrices, corresponding respectively to the key, query, value and output matrices. Note that other works extend this approach to the feed-forward module's weight matrices~\cite{he2021towards}.
\paragraph{LoRA for VLMs.} A straightforward way to apply LoRA in vision-language is to apply it to all the matrices of the vision and text encoders. However, due to the relatively small supervision inherent to the few-shot setting, we only apply low-rank matrices on the query, key and value matrices with $r=2$. We regularize the input of the LoRA module by a dropout layer with $p=0.25$~\cite{hu2021lora}. The number of iterations is set equal to 500 times N/K (the number of labeled samples per class). We used a learning rate of $2*10^{-4}$, with a cosine scheduler and a batch size of 32, so that all training could be performed on a single GPU of 24 Gb. {\em These hyper-parameters are kept fixed across all the experiments}. The input prompt is simply set to \texttt{a photo of a} [$\texttt{kth class name}$], $k=1,\dots, K$, for every dataset, to emphasize the applicability of CLIP-LoRA without resorting to complex initial manual prompting. 
Note that the LoRA modules are positioned at every levels of both encoders. The impact of the location of the LoRA modules is studied in Section \ref{sec:ablation}, putting in evidence that adapting both modalities can be necessary for certain tasks.

%% file: text/4_results.tex
\input{tables/few_shot_bis}
\section{Few-shot learning}
\input{figures/few_shot}

We follow the setting of previous work~\cite{coop}. We consider 10 datasets for fine-grained classification of scenes (SUN397~\cite{sun397}), aircraft types (Aircraft~\cite{aircraft}), satellite imagery (EuroSAT~\cite{eurosat}), automobiles (StanfordCars~\cite{cars}), food items (Food101~\cite{food}), pet breeds (OxfordPets~\cite{pets}), flowers (Flower102~\cite{flower}), general objects (Caltech101~\cite{caltech101}), textures (DTD~\cite{dtd}) and human actions (UCF101~\cite{ucf101}) as well as ImageNet~\cite{imagenet}.
These datasets offer a thorough benchmarking framework for evaluating few-shot visual classification tasks.
\paragraph{Comparative methods.} We compare CLIP-LoRA to several prompt-based methods: CoOp~\cite{coop} (4) with 4 learnable tokens, CoOp~\cite{coop} (16) with 16 learnable tokens, CoCoOp~\cite{cocoop}, PLOT++~\cite{plot} which is an adaption of the original PLOT proposed by the same authors specifically designed for transformer architectures, KgCoOp~\cite{kgcoop}, MaPLe~\cite{khattak2023maple} for which we follow the training procedure of their "base-to-new" setting, ProGrad~\cite{prograd} with 16 tokens. We also report adapter-based methods: Tip-Adapter-F~\cite{tip-adapter} for which we reduce the validation set to a reasonable size of min(n\_shots, 4), TaskRes~\cite{yu2023task} for which we only report the not enhanced base performance due to its unavailability for all datasets/shots/backbones studied in this paper. Despite some questionable arbitrary choices as discussed in~\cite{Julio-ArXiv23}, we keep their specific hyper-parameters~\cite{Julio-ArXiv23} while CLIP-LoRA uses the same hyper-parameters for every tasks.
\input{figures/ablation}
\paragraph{CLIP-LoRA outperforms, on average, adapter- and prompt-based few-shot methods.} The strongest adapter-based method in Table \ref{tab:few_shot_datasets} is Tip-Adapter-F, which is not competitive with CLIP-LoRA despite relying heavily on arbitrary hyper-parameters for each dataset (namely the starting value of their $\alpha, \beta$ as well as the search range during validation). We can conclude the same for TaskRes, which also relies on arbitrary choices for a given dataset, i.e., a specific learning rate for ImageNet and a specific scaling factor for the Flowers dataset. Regarding prompt-based approaches, Table \ref{tab:few_shot_datasets} shows that CoOp and ProGrad are outperformed by a large margin. The strongest competitor is, without a doubt, PLOT++. PLOT++ necessitates a two-stage training (each of 50 epochs for ImageNet) as well as several dataset-specific textual templates for their optimal transport formulation, reducing its portability to other downstream tasks. Overall, CLIP-LoRA performs better, especially on ImageNet, UCF101 and Aircraft, while being more practical. However, it underperforms on two datasets, Food101 and OxfordPet, where few-shot learning offers minimal improvement. This may be attributed to the lack of regularization, considering we use straightforward cross-entropy loss. We observe a similar trend with CoOp, whereas approaches that incorporate explicit regularization, such as ProGrad, do not exhibit this issue. Note that more detailed results, including for 2 and 8 shots, are available in the Appendix.
\paragraph{CLIP-LoRA performances are consistent across various vision encoders.} As depicted in Figure \ref{fig:few_shots}, CLIP-LoRA surpasses, on average, the other few-shot methods with both the ViT-B/32 architecture and the larger ViT-L/14. This further supports the versatility of our approach. Detailed results for the three backbones are available in the Appendix.
\paragraph{CLIP-LoRA is computationally and memory efficient.} Table \ref{tab:runtime} compares the training time of the leading prompt-learning methods; CLIP-LoRA achieves better performance with shorter training. Moreover, the best performing adapter method,  namely Tip-Adapter-F, depends on a large cache model that stores embeddings for all instances across every class. In contrast, LoRA merges its adapted matrices at the inference stage, thereby eliminating the need for extra memory beyond what is required by the original model.

%% file: tables/few_shot_bis.tex
\begin{table*}[t!]
\caption{Detailed results for 11 datasets with the ViT-B/16 as visual backbone. Top-1 accuracy averaged over 3 random seeds is reported. Highest value is highlighted in \textbf{bold}, and the second highest is \underline{underlined}.}
\label{tab:few_shot_datasets}
\centering
\resizebox{\textwidth}{!}{
\renewcommand{\arraystretch}{1.06} 
\begin{tabular}{llcccccccccccc}
\toprule
Shots & Method & ImageNet & SUN & Aircraft & EuroSAT & Cars & Food & Pets &  Flowers & Caltech & DTD & UCF & Average
\\ \midrule 
\multirow{1}{*}{0} & \textbf{CLIP} {\tiny \textbf{(ICML '21)}} & 66.7 & 62.6 & 24.7 & 47.5 & 65.3 & 86.1 & 89.1 & 71.4 & 92.9 & 43.6 & 66.7 & 65.1 \\
\midrule
\midrule
\multirow{11}{*}{\textbf{1}} 
& CoOp (4) {\tiny \textbf{(IJCV '22)}} & 68.0 & 67.3 & 26.2 & 50.9 & 67.1 & 82.6 & 90.3 & 72.7 & 93.2 & 50.1 & 70.7 & 67.2 \\
& CoOp (16) {\tiny \textbf{(IJCV '22)}} & 65.7 & 67.0 & 20.8 & 56.4 & 67.5 & 84.3 & 90.2 & 78.3 & 92.5 & 50.1 & 71.2 & 67.6 \\ 
& CoCoOp {\tiny \textbf{(CVPR '22)}} & 69.4 & 68.7 & 28.1 & 55.4 & 67.6 & 84.9 & 91.9 & 73.4 & 94.1 & 52.6 & 70.4 & 68.8 \\
& TIP-Adapter-F {\tiny \textbf{(ECCV '22)}} & 69.4 & 67.2 & 28.8 & \underline{67.8} & 67.1 & 85.8 & 90.6 & \textbf{83.8} & 94.0 & 51.6 & 73.4 & \underline{70.9} \\
& CLIP-Adapter {\tiny \textbf{(IJCV '23)}} & 67.9 & 65.4 & 25.2 & 49.3 & 65.7 & 86.1 & 89.0 & 71.3 & 92.0 & 44.2 & 66.9 & 65.7 \\
& PLOT++ {\tiny \textbf{(ICLR '23)}} & 66.5 & 66.8 & 28.6 & 65.4 & \underline{68.8} & \underline{86.2} & 91.9 & 80.5 & \textbf{94.3} & \textbf{54.6} & \underline{74.3} & 70.7 \\
& KgCoOp {\tiny \textbf{(CVPR '23)}} & 68.9 & 68.4 & 26.8 & 61.9 & 66.7 & \textbf{86.4} & \underline{92.1} & 74.7 & \underline{94.2} & 52.7 & 72.8 & 69.6 \\
& TaskRes {\tiny \textbf{(CVPR '23)}} & 69.6 & 68.1 & \textbf{31.3} & 65.4 & \underline{68.8} & 84.6 & 90.2 & 81.7 & 93.6 & 53.8 & 71.7 & 70.8 \\
& MaPLe {\tiny \textbf{(CVPR '23)}} & \underline{69.7} & \underline{69.3} & 28.1 & 29.1 & 67.6 & 85.4 & 91.4 & 74.9 & 93.6 & 50.0 & 71.1 & 66.4 \\
& ProGrad {\tiny \textbf{(ICCV '23)}} & 67.0 & 67.0 & 28.8 & 57.0 & 68.2 & 84.9 & 91.4 & 80.9 & 93.5 & 52.8 & 73.3 & 69.5  \\
\rowcolor{LightGray} & CLIP-LoRA (Ours) & \textbf{70.4} & \textbf{70.4} & \underline{30.2} & \textbf{72.3} & \textbf{70.1} & 84.3 & \textbf{92.3} & \underline{83.2} & 93.7 & \underline{54.3} & \textbf{76.3} & \textbf{72.5} \\
\midrule
\midrule
\multirow{11}{*}{\textbf{4}}
& CoOp (4)  {\tiny \textbf{(IJCV '22)}} &  69.7 & 70.6 & 29.7 & 65.8 & 73.4 & 83.5 & 92.3 & 86.6 & 94.5 & 58.5 & 78.1 & 73.0 \\
& CoOp (16)  {\tiny \textbf{(IJCV '22)}} & 68.8 & 69.7 & 30.9 & 69.7 & 74.4 & 84.5 & 92.5  & 92.2 & 94.5 & 59.5 & 77.6 & 74.0 \\
& CoCoOp {\tiny \textbf{(CVPR '22)}} & 70.6 & 70.4 & 30.6 & 61.7 & 69.5 & 86.3 & \underline{92.7} & 81.5 & 94.8 & 55.7 & 75.3 & 71.7 \\
& TIP-Adapter-F {\tiny \textbf{(ECCV '22)}} & 70.7 & 70.8 & \underline{35.7} & 76.8 & 74.1 & 86.5 & 91.9 & 92.1 & 94.8 & 59.8 & 78.1 & 75.6 \\
& CLIP-Adapter {\tiny \textbf{(IJCV '23)}} & 68.6 & 68.0 & 27.9 & 51.2 & 67.5 & 86.5 & 90.8 & 73.1 & 94.0 & 46.1 & 70.6 & 67.7 \\
& PLOT++ {\tiny \textbf{(ICLR '23)}} & 70.4 & 71.7 & 35.3 & \underline{83.2} & \underline{76.3} & 86.5 & 92.6 & \underline{92.9} & \underline{95.1} & \underline{62.4} & \underline{79.8} & \underline{76.9} \\
& KgCoOp {\tiny \textbf{(CVPR '23)}} & 69.9 & 71.5 & 32.2 & 71.8 & 69.5 & \textbf{86.9} & 92.6 & 87.0 & 95.0 & 58.7 & 77.6 & 73.9 \\
& TaskRes {\tiny \textbf{(CVPR '23)}} & \underline{71.0} & \underline{72.7} & 33.4 & 74.2 & 76.0 & 86.0 & 91.9 & 85.0 & 95.0 & 60.1 & 76.2 & 74.7 \\
& MaPLe {\tiny \textbf{(CVPR '23)}} & 70.6 & 71.4 & 30.1 & 69.9 & 70.1 & \underline{86.7} & \textbf{93.3} & 84.9 & 95.0 & 59.0 & 77.1 & 73.5  \\
& ProGrad {\tiny \textbf{(ICCV '23)}} & 70.2 & 71.7 & 34.1 & 69.6 & 75.0 & 85.4 & 92.1 & 91.1 & 94.4 & 59.7 & 77.9 & 74.7 \\
\rowcolor{LightGray} & CLIP-LoRA (Ours) & \textbf{71.4} & \textbf{72.8} & \textbf{37.9} & \textbf{84.9} & \textbf{77.4} & 82.7 & 91.0 & \textbf{93.7} & \textbf{95.2} & \textbf{63.8} & \textbf{81.1} & \textbf{77.4} \\
\midrule
\midrule
\multirow{11}{*}{\textbf{16}}
& CoOp (4)  {\tiny \textbf{(IJCV '22)}} &  71.5 & 74.6 & 40.1 & 83.5 & 79.1 & 85.1 & 92.4 & 96.4 & 95.5 & 69.2 & 81.9 & 79.0 \\
& CoOp (16) {\tiny \textbf{(IJCV '22)}} & 71.9 & 74.9 & 43.2 & 85.0 & 82.9 & 84.2 & 92.0 & 96.8 & 95.8 & 69.7 & 83.1 & 80.0 \\
& CoCoOp {\tiny \textbf{(CVPR '22)}} & 71.1 & 72.6 & 33.3 & 73.6 & 72.3 & \textbf{87.4} & \underline{93.4} & 89.1 & 95.1 & 63.7 & 77.2 & 75.4 \\
& TIP-Adapter-F {\tiny \textbf{(ECCV '22)}} & \underline{73.4} &  \underline{76.0} & 44.6 & 85.9 & 82.3 & 86.8 & 92.6 & 96.2 & 95.7 & 70.8 & 83.9 & 80.7 \\
& CLIP-Adapter {\tiny \textbf{(IJCV '23)}} & 69.8 & 74.2 & 34.2 & 71.4 & 74.0 & 87.1 & 92.3 & 92.9 & 94.9 & 59.4 & 80.2 & 75.5  \\
& PLOT++ {\tiny \textbf{(ICLR '23)}} & 72.6 & \underline{76.0} & \underline{46.7} & \underline{92.0} & \underline{84.6} & 87.1 & \textbf{93.6} & \underline{97.6} & \underline{96.0} & 71.4 & \underline{85.3} & \underline{82.1} \\
& KgCoOp {\tiny \textbf{(CVPR '23)}} & 70.4 & 73.3 & 36.5 & 76.2 & 74.8 & \underline{87.2} & 93.2 & 93.4 & 95.2 & 68.7 & 81.7 & 77.3 \\
& TaskRes {\tiny \textbf{(CVPR '23)}} & 73.0 & \textbf{76.1} & 44.9 & 82.7 & 83.5 & 86.9 & 92.4 & 97.5 & 95.8 & \underline{71.5} & 84.0 & 80.8 \\
& MaPLe {\tiny \textbf{(CVPR '23)}} &  71.9 & 74.5 & 36.8 & 87.5 & 74.3 & \textbf{87.4} & 93.2 & 94.2 & 95.4 & 68.4 & 81.4 & 78.6 \\
& ProGrad {\tiny \textbf{(ICCV '23)}} &  72.1 & 75.1 & 43.0 & 83.6 & 82.9 & 85.8 & 92.8 & 96.6 & 95.9 & 68.8 & 82.7 & 79.9 \\
\rowcolor{LightGray} & CLIP-LoRA (Ours) & \textbf{73.6} & \textbf{76.1} & \textbf{54.7} & \textbf{92.1}  & \textbf{86.3} & 84.2 & 92.4 & \textbf{98.0} & \textbf{96.4} & \textbf{72.0} & \textbf{86.7} & \textbf{83.0} \\
\bottomrule

\end{tabular}}

\end{table*}

%% file: figures/few_shot.tex

\begin{figure*}[t]
\centering
\subfloat{\label{sfig:a}\includegraphics[width=.2\textwidth]{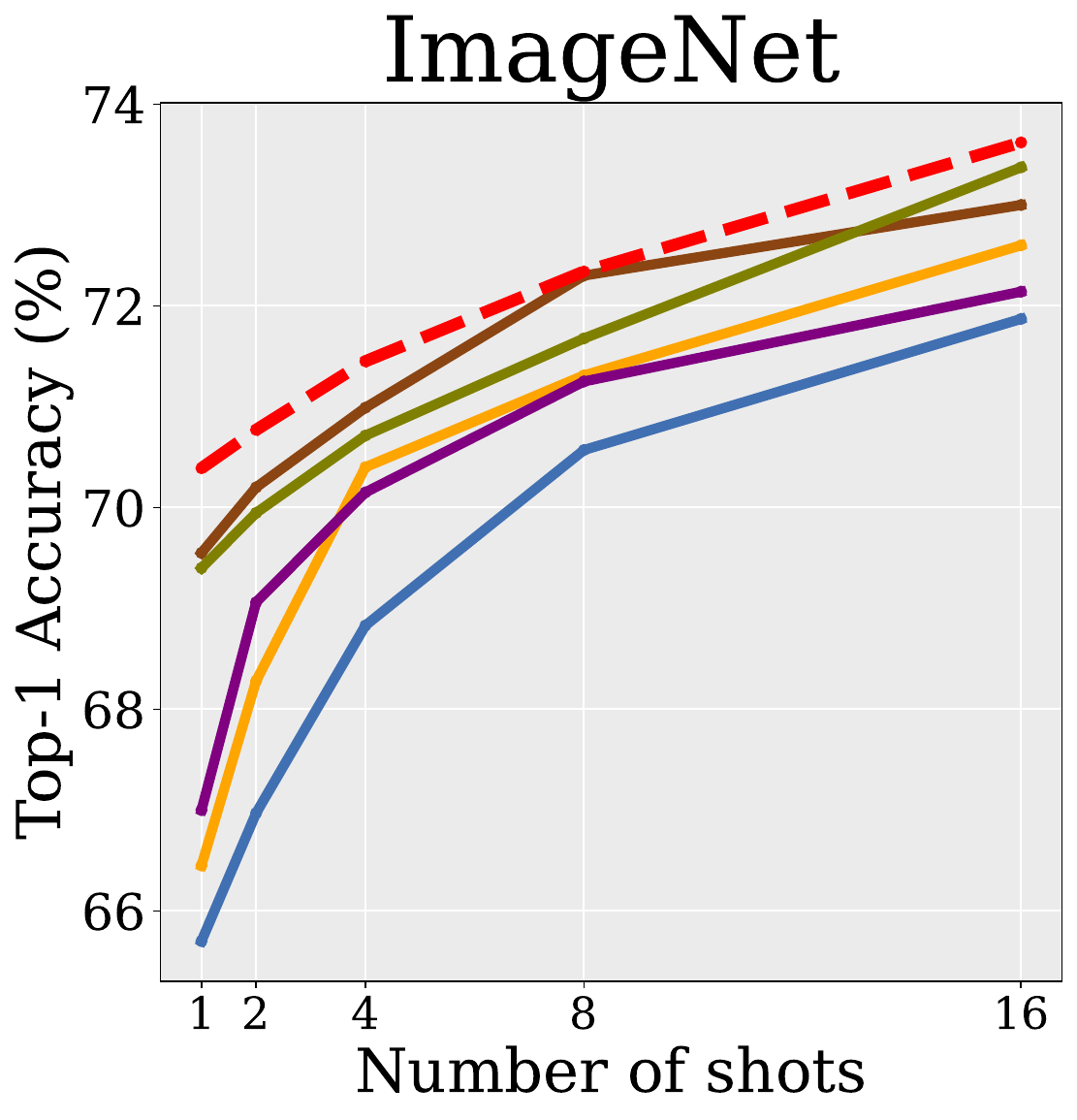}}\hfill
\subfloat{\label{sfig:b}\includegraphics[width=.2\textwidth]{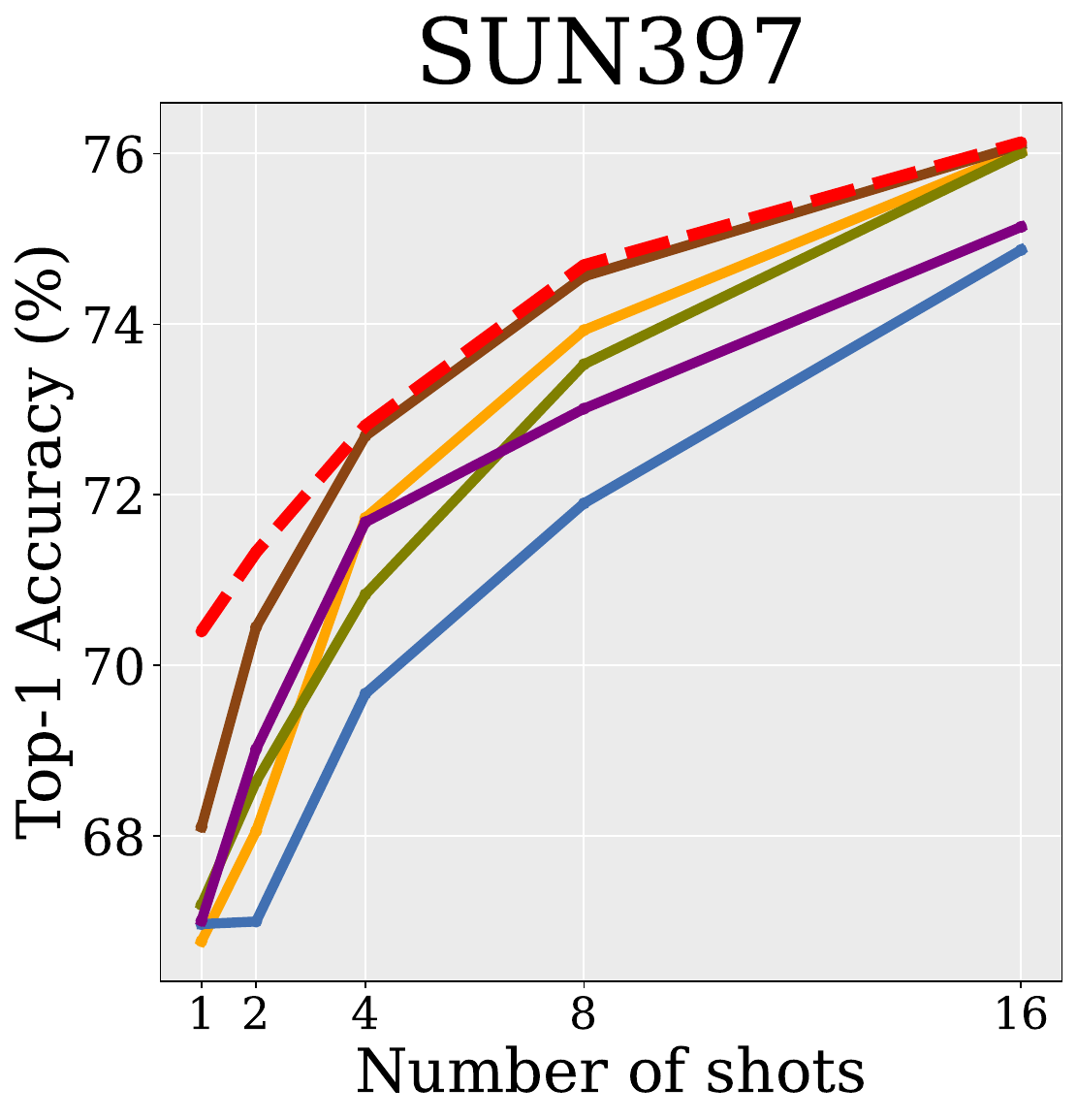}}\hfill
\subfloat{\label{sfig:h}\includegraphics[width=.2\textwidth]{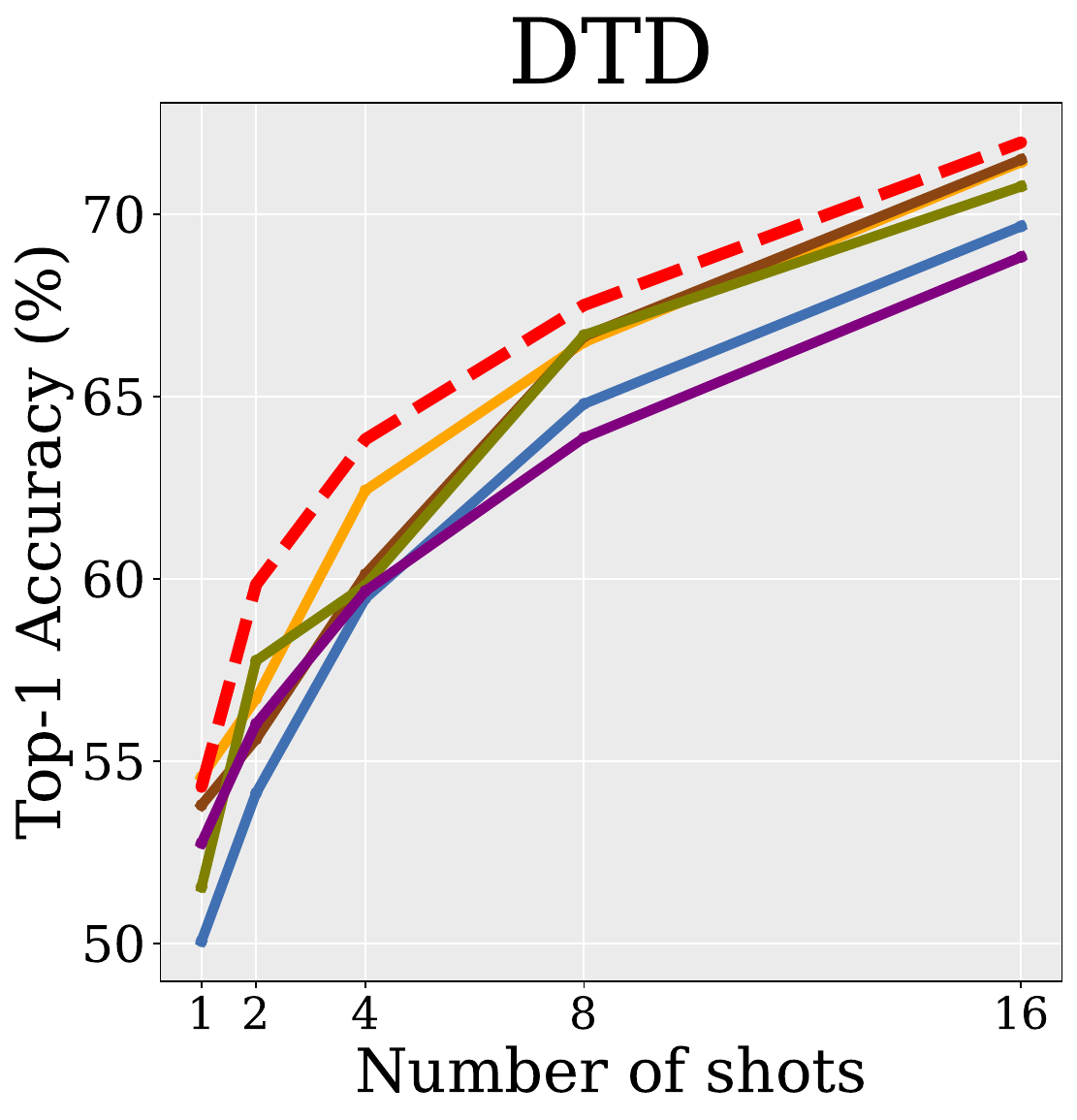}}\hfill
\subfloat{\label{sfig:f}\includegraphics[width=.2\textwidth]{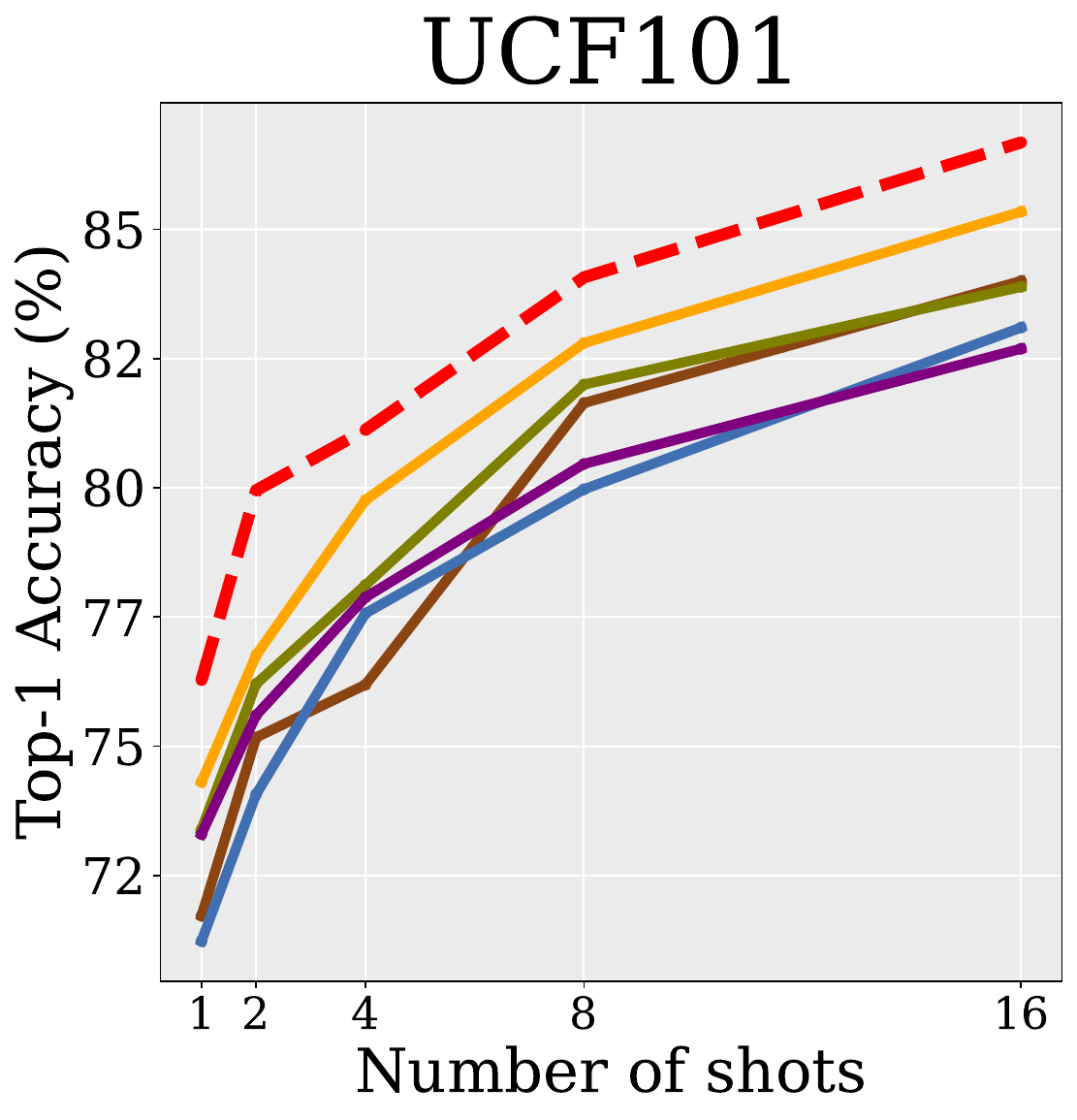}}\hfill
\subfloat{\label{sfig:c}\includegraphics[width=.2\textwidth]{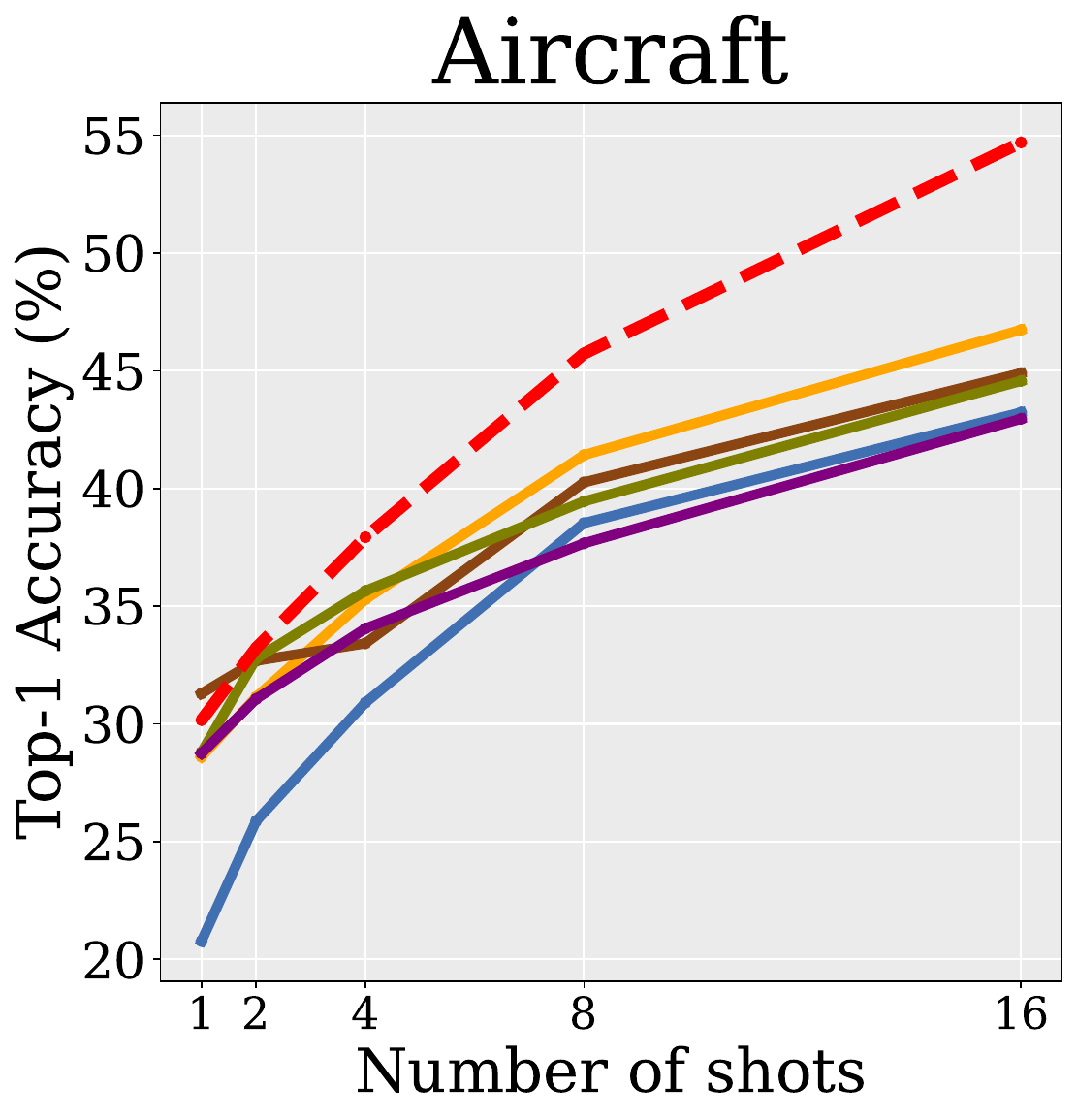}}\hfill
\subfloat{\label{sfig:d}\includegraphics[width=.2\textwidth]{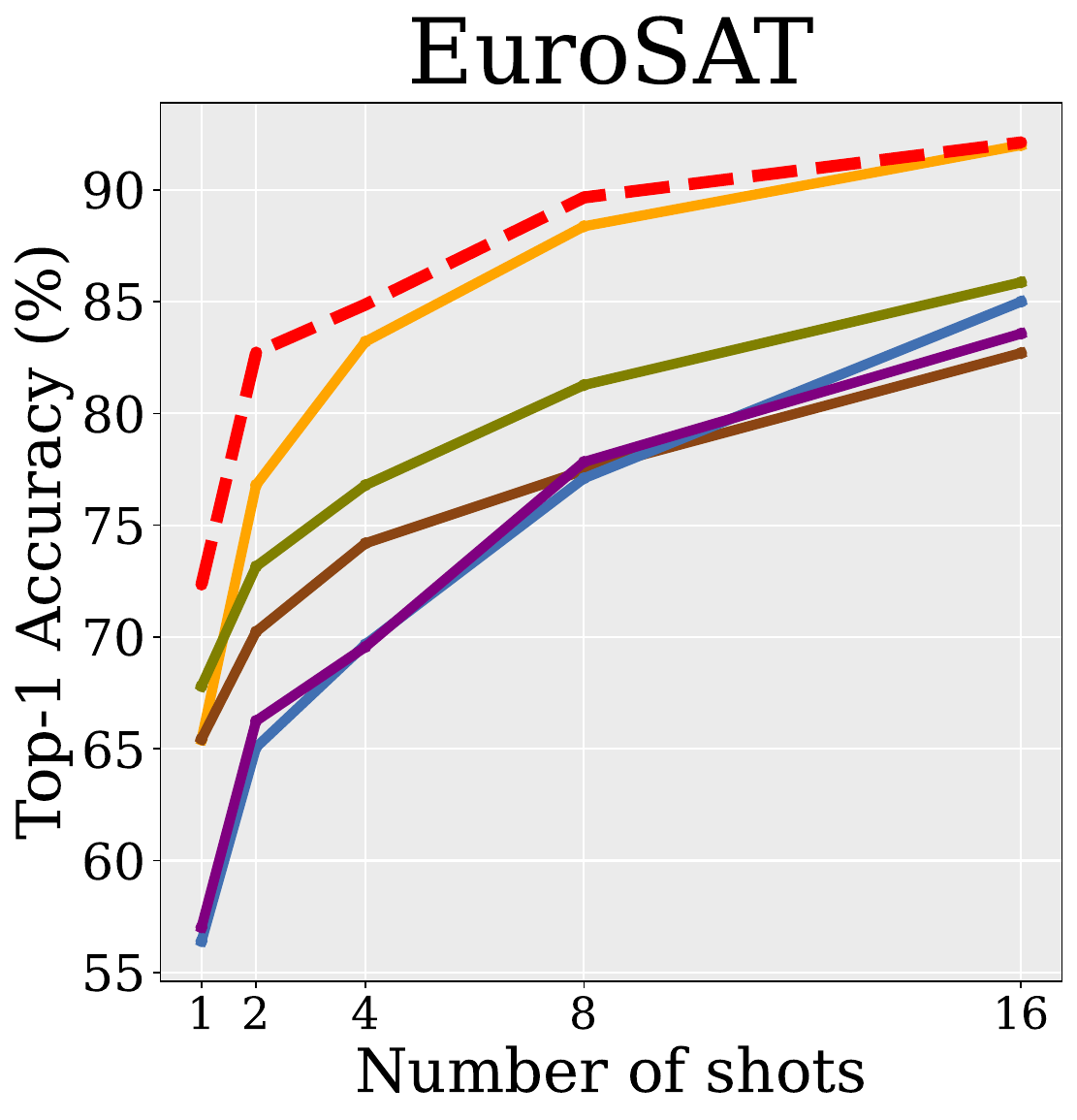}}\hfill
\subfloat{\label{sfig:f}\includegraphics[width=.2\textwidth]{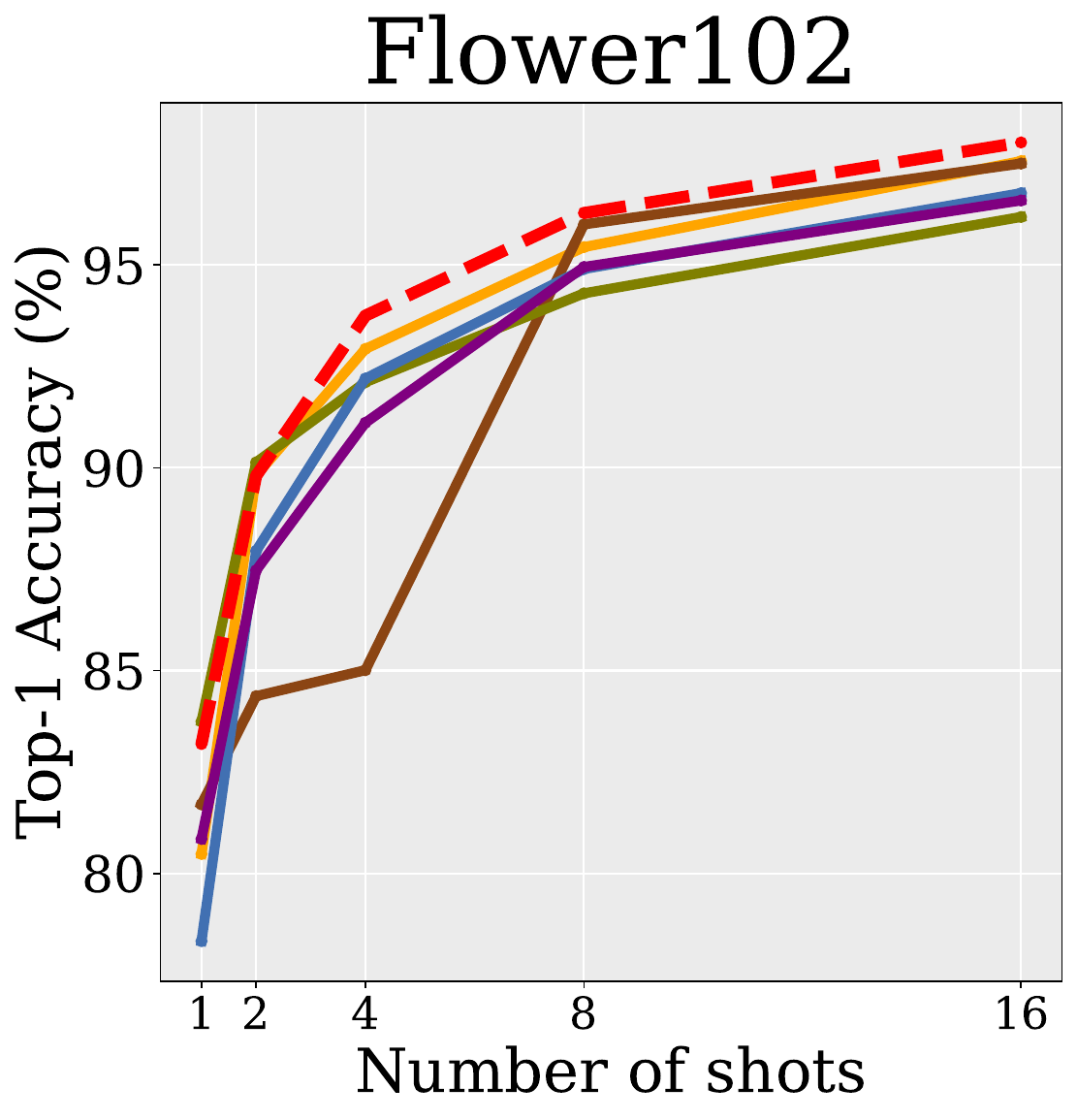}}\hfill
\subfloat{\label{sfig:g}\includegraphics[width=.2\textwidth]{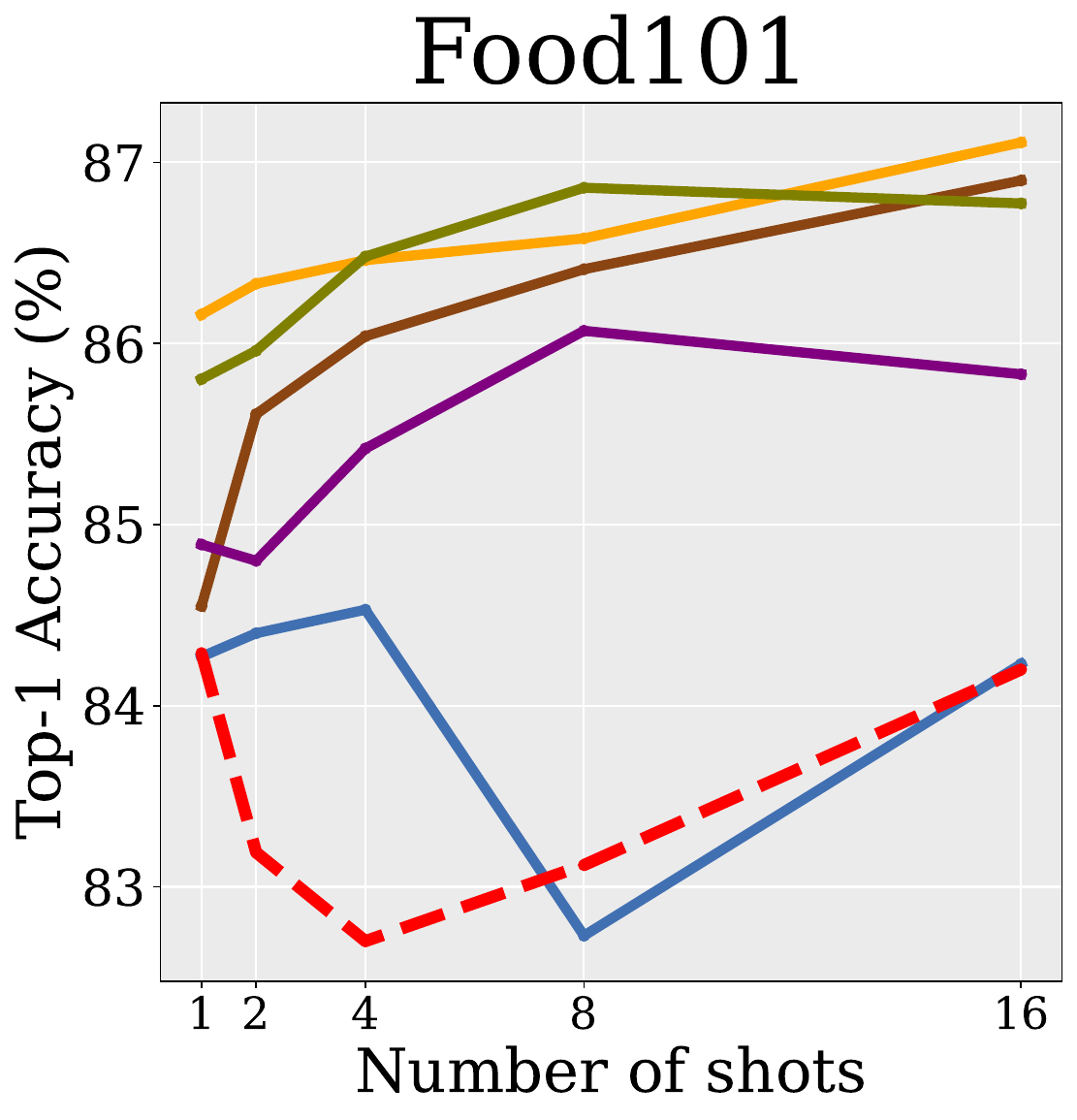}}\hfill
\subfloat{\label{sfig:h}\includegraphics[width=.2\textwidth]{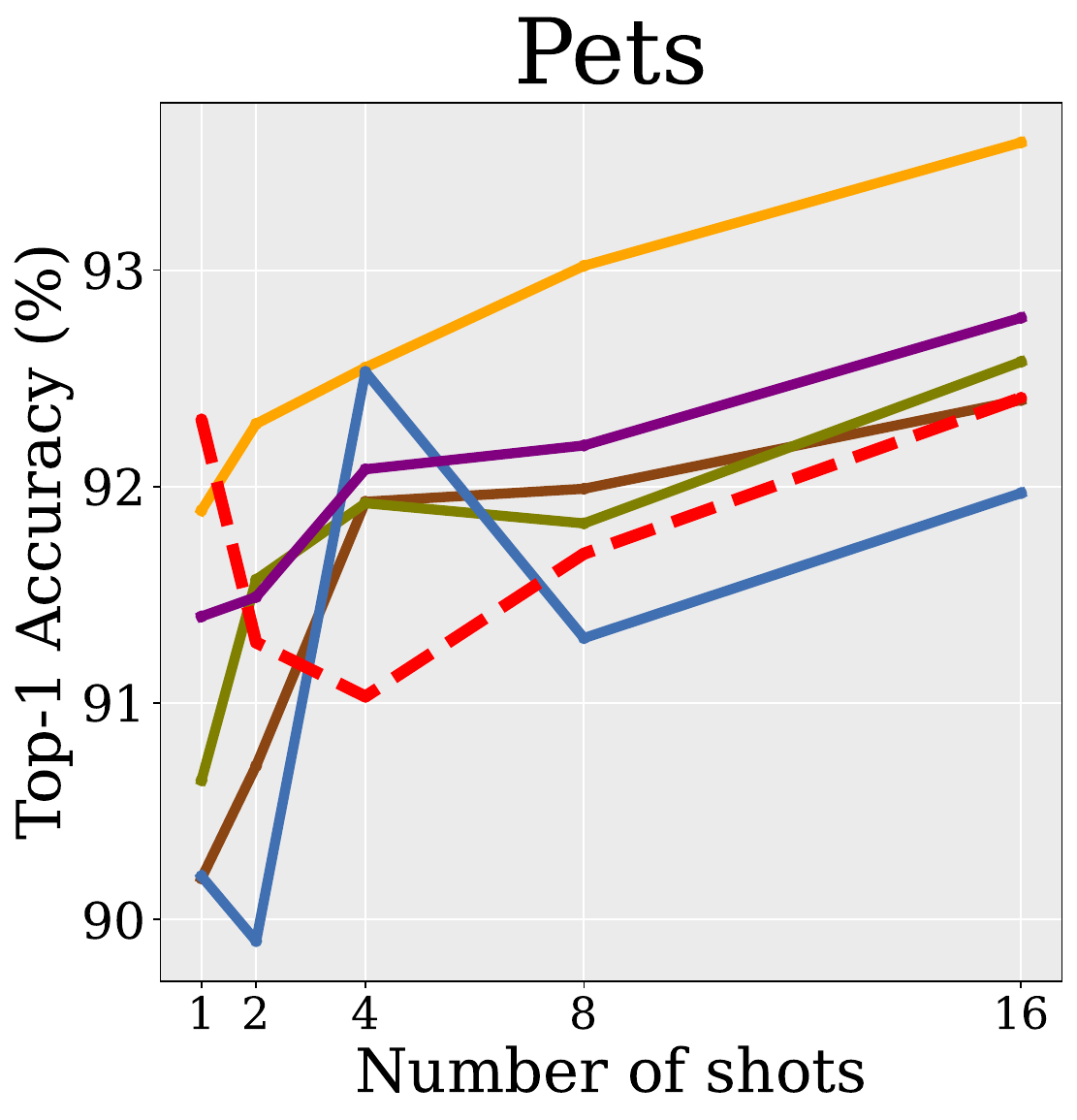}}\hfill
\subfloat{\label{sfig:i}\includegraphics[width=.2\textwidth]{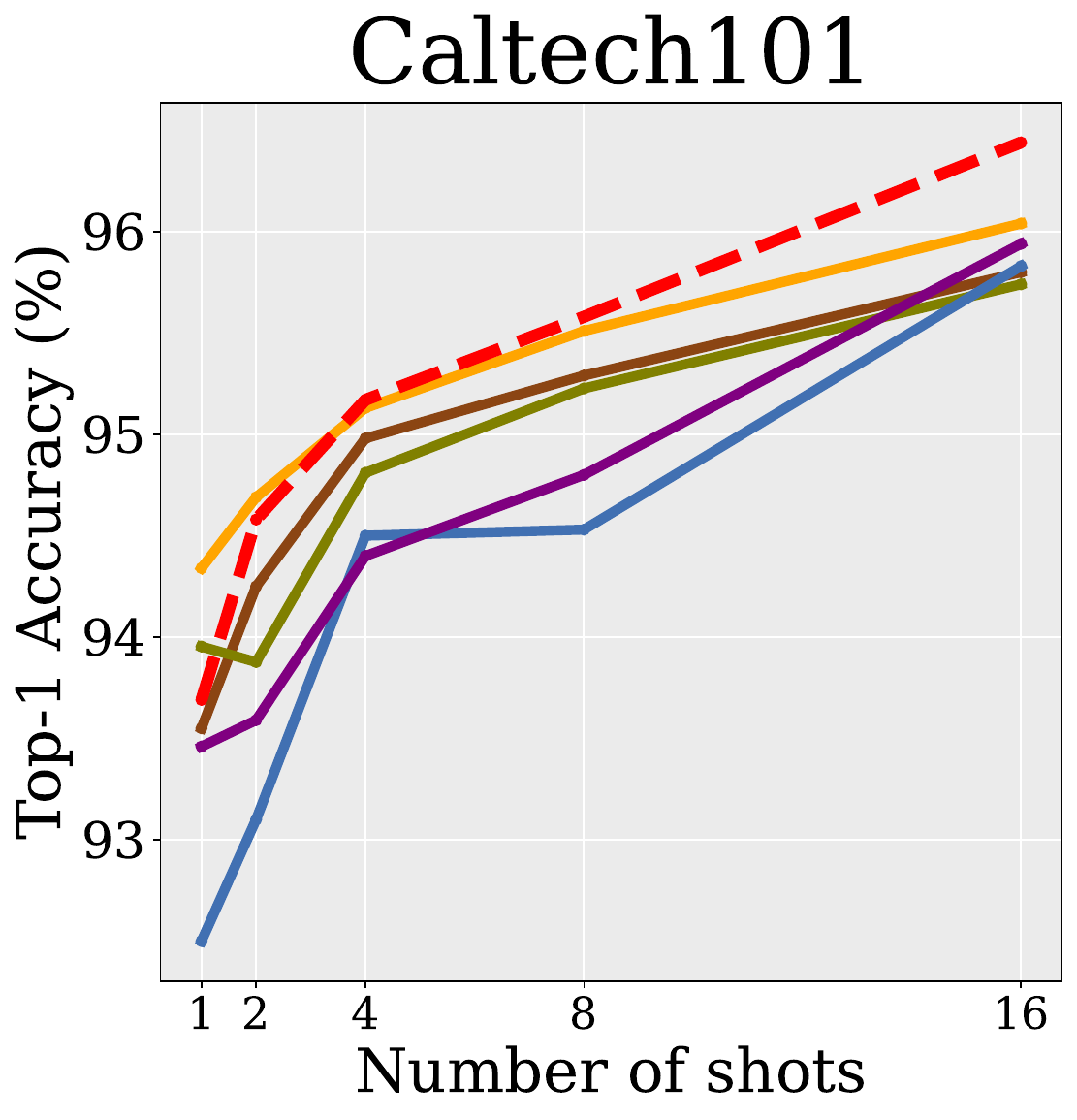}}\hfill
\subfloat{\label{sfig:g}\includegraphics[width=.2\textwidth]{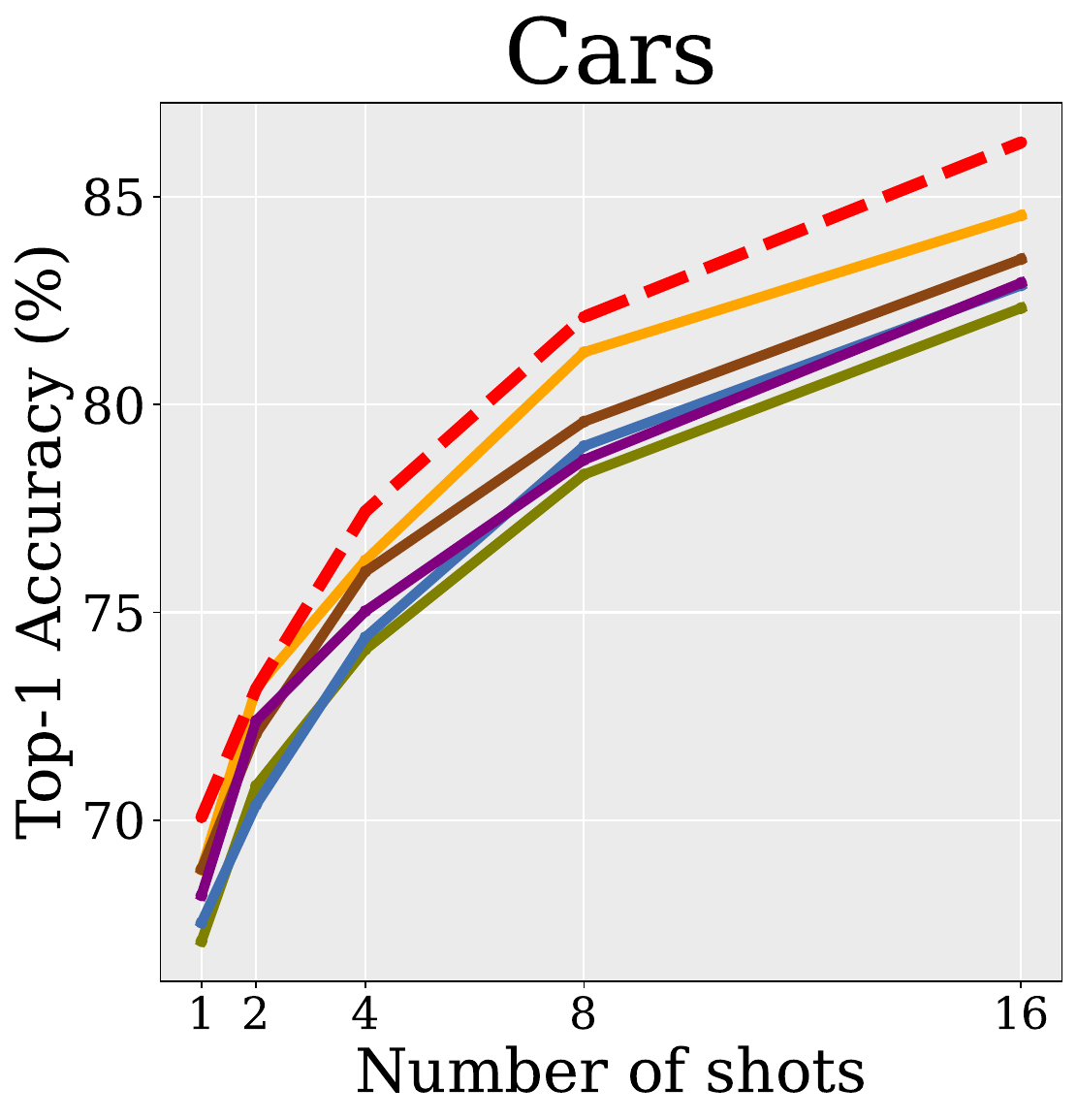}}\hfill
\subfloat{\label{sfig:h}\includegraphics[width=.2\textwidth]{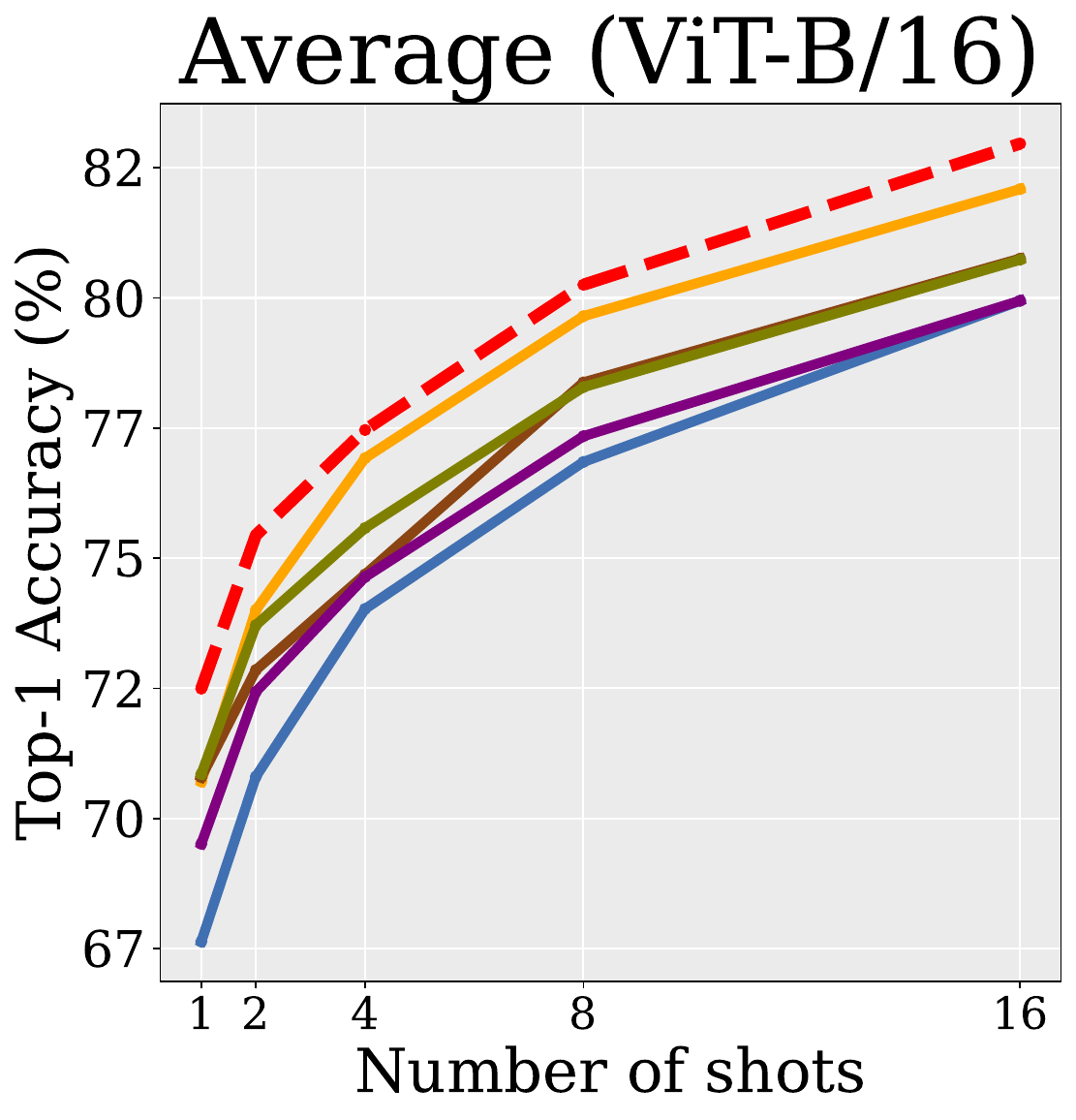}}\hfill
\subfloat{\label{sfig:i}\includegraphics[width=.2\textwidth]{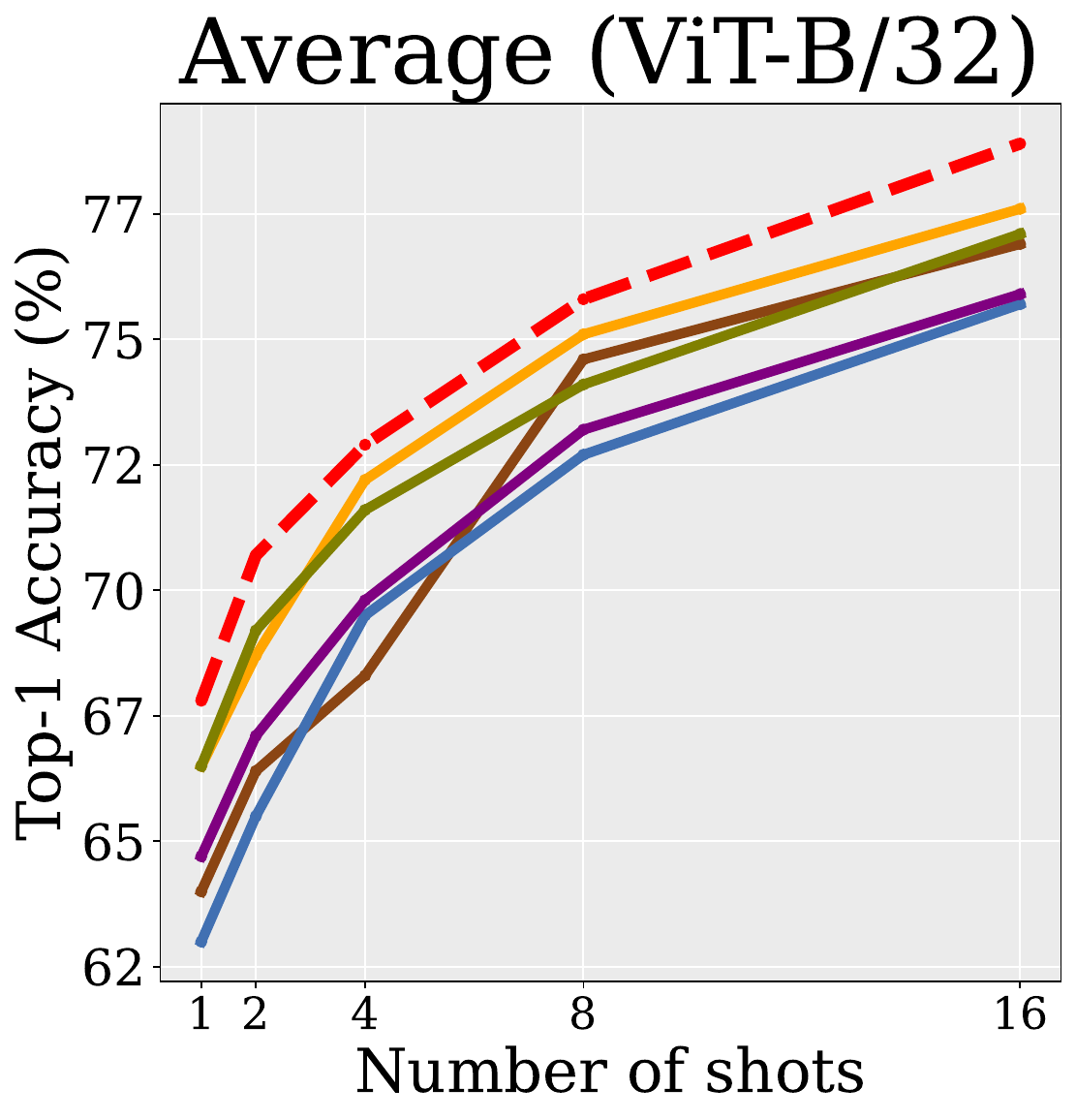}}\hfill
\subfloat{\label{sfig:g}\includegraphics[width=.2\textwidth]{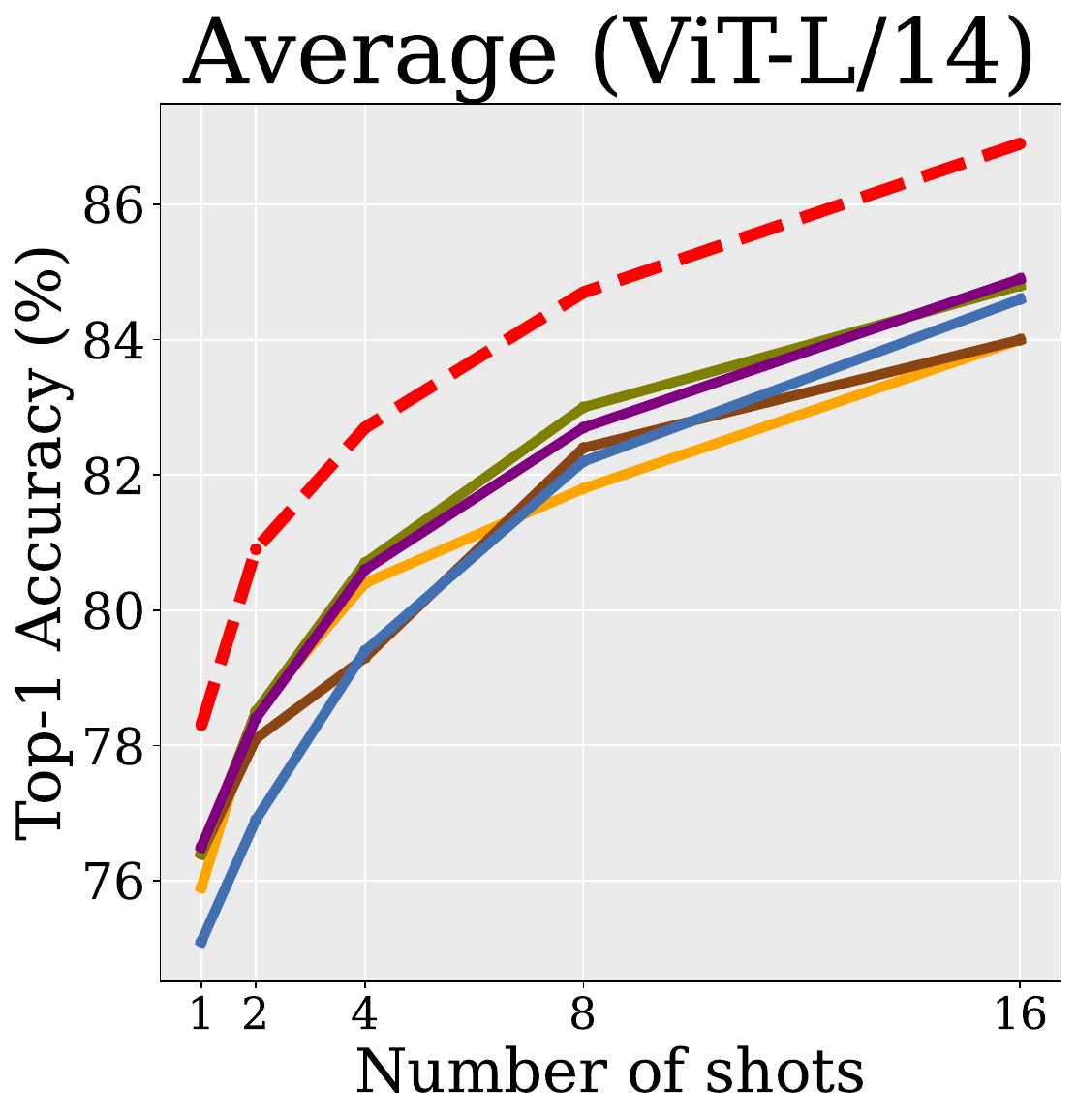}}\hfill
\subfloat{\label{sfig:i}\includegraphics[width=.2\textwidth]{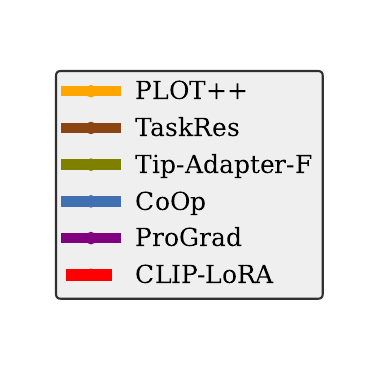}}\hfill

\caption{Detailed few-shot learning results on the 10 fine-grained datasets and ImageNet with the ViT-B/16 visual backbone. Average performance for the ViT-B/16, ViT-B/32 and ViT-L/14 on the same 11 datasets is reported in the last three plots, respectively.}
\label{fig:few_shots}
\end{figure*}

%% file: figures/ablation.tex
\begin{figure*}[t]
\centering
\begin{subfigure}[b]{\textwidth}
    \centering
   \includegraphics[width=0.96\linewidth]{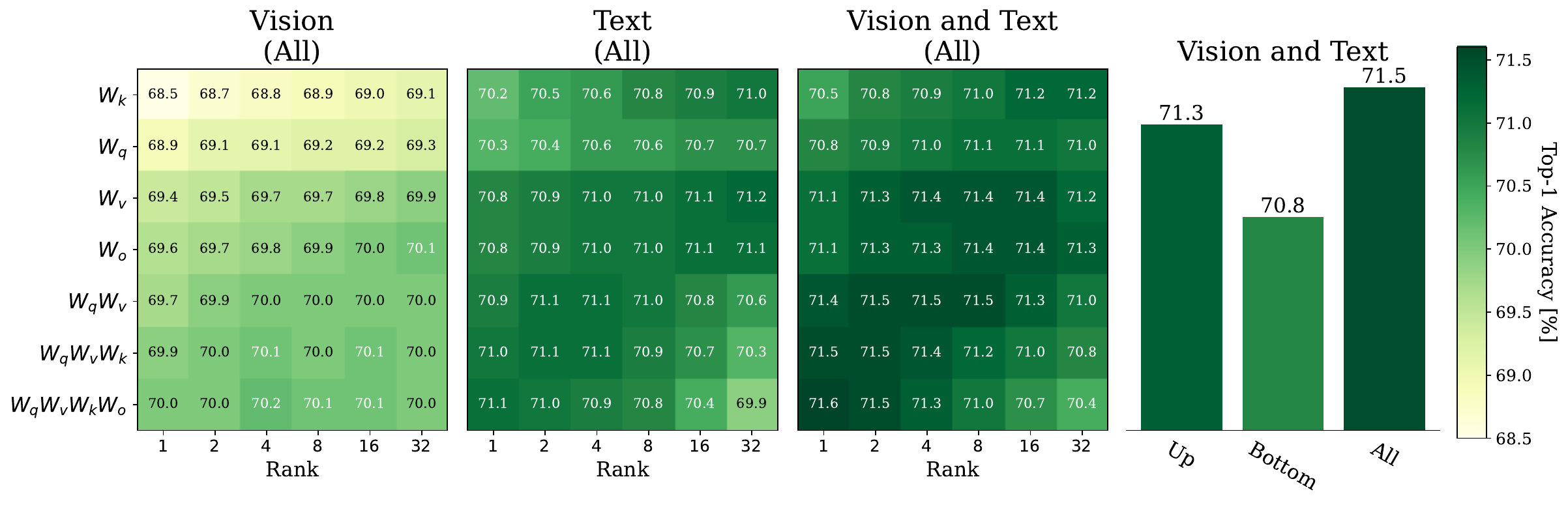}
   \caption{ImageNet}
   \label{fig:ablation_imagenet} 
\end{subfigure}

\begin{subfigure}[b]{\textwidth}
    \centering
   \includegraphics[width=0.96\linewidth]{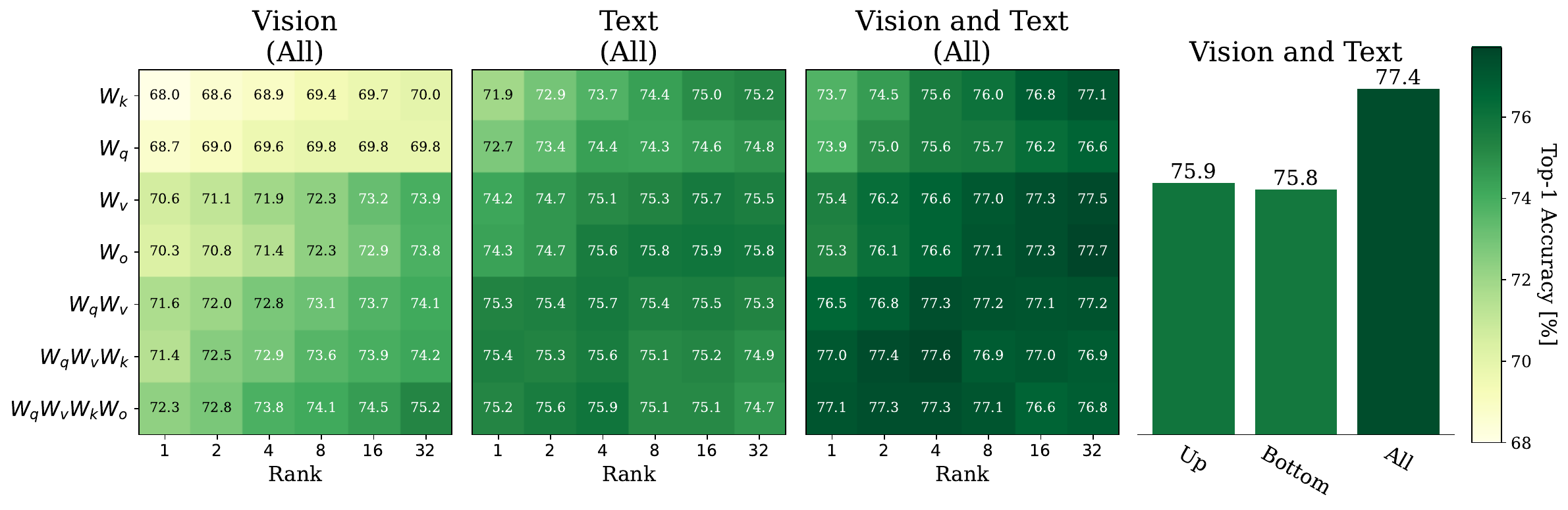}
   \caption{Stanford Cars}
   \label{fig:ablation_cars}
\end{subfigure}

\begin{subfigure}[b]{\textwidth}
    \centering
   \includegraphics[width=0.96\linewidth]{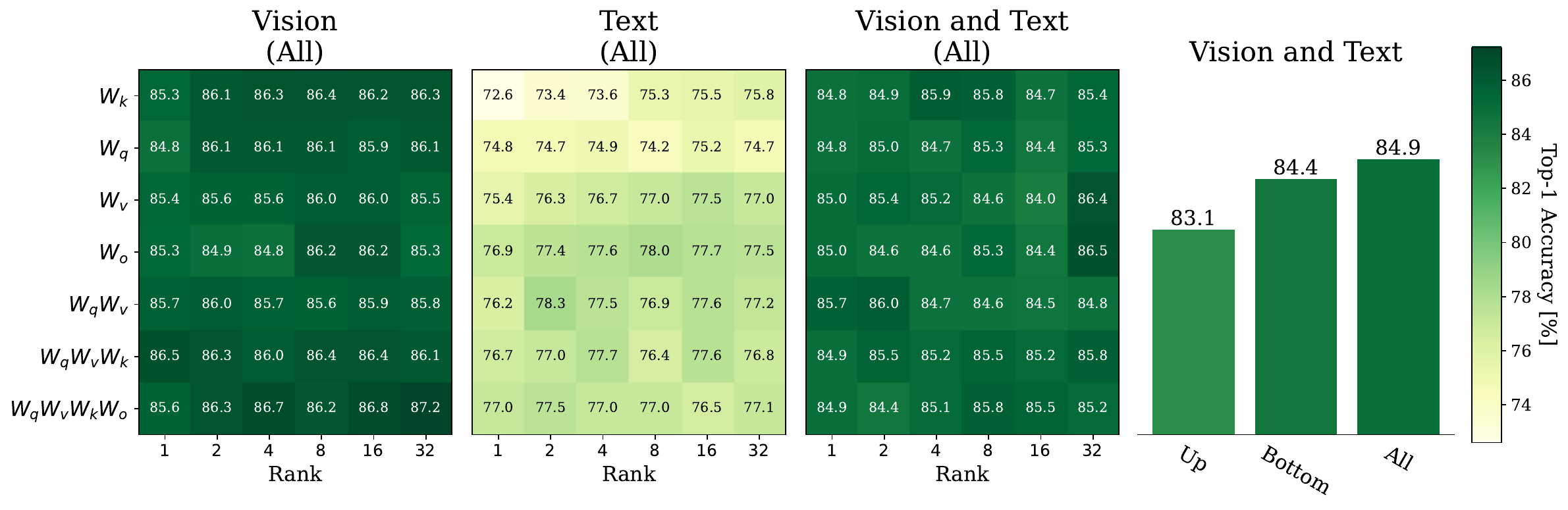}
   \caption{EuroSAT}
   \label{fig:ablation_eurosat}
\end{subfigure}

\caption{Top-1 accuracy with 4-shots for different matrices of the attention bloc and increasing rank, when the low-rank matrices are positioned at every level of the encoders (All). The fourth bar plot study the impact of positioning the low-rank matrices only on the half last levels (Up), the first half levels (Bottom), or at every level (All). Reported top-1 accuracy is averaged over 3 random seeds.}
\label{fig:ablation}
\end{figure*}

%% file: text/5_ablation.tex
\section{How to apply LoRA for VLMs?}
\label{sec:ablation}
In this section, we delve into the utilization of LoRA modules, identifying three principal design considerations: (1) the choice between tuning the vision encoder, the text encoder, or both, including the specific layers to adjust; (2) the selection of attention matrices for tuning; and (3) the determination of the appropriate rank for these matrices. We explore these aspects across three datasets: ImageNet, Stanford Cars, and EuroSAT. ImageNet was selected for its broad diversity, while the latter two were chosen for their distinctive behaviors. Results are depicted in Figure \ref{fig:ablation} for seven different groups of adapted attention matrices and increasing rank value.
\paragraph{Adapting both encoders leads to the best results on average.} With the exception of EuroSAT, where adapting solely the vision encoder shows marginally better stability, tuning both encoders concurrently is the most effective strategy, leading to significant enhancements. This aligns with recent approaches that incorporate additional vision tokens~\cite{plot, khattak2023maple} to augment performance beyond what is achievable with text-only prompt tuning, as seen in CoOp~\cite{coop}.
\paragraph{Tuning more attention matrices can lead to better results but...} Among the four attention matrices studied, adapting value or output matrices ($\mathbf W_{v}$ and $\mathbf W_{o}$) appears to be the best strategy, showing quite consistent differences in performance. Moreover, as discussed in the original LoRA paper and subsequent works~\cite{hu2021lora, zhang2022adaptive}, adapting a larger number of weight matrices can lead to better results. However, it can also decrease performance, as demonstrated on ImageNet and StanfordCars with high rank. This is in line with recent methods that aim to dynamically adjust the rank of the matrices~\cite{zhang2022adaptive, valipour2022dylora}. 
\paragraph{Choosing the location of LoRA modules requires careful consideration.} The impact of LoRA module placement—whether on the lower half (bottom), or the upper half (up)—is illustrated in the bar plots of Figure \ref{fig:ablation} with varying performance and no clear winner. We found it more effective to add LoRA modules across all layers. In comparison, in the context of LLMs, AdaLoRA~\cite{zhang2022adaptive} suggests that allocating a larger rank to the middle and last layers rather than the first ones yields better results. Similar strategies applied for VLMs could reveal promising avenues for future research.

%% file: text/6_conclusion.tex
\section{Conclusion}
We established a strong baseline by consistently outperforming prompt- and adapter-based methods in few-shot adaptation of Vision-Language Models (VLMs) using fixed hyper-parameters. We hope our work inspires future efforts to design methods that either uphold this simplicity and efficiency with fixed hyper-parameters or offer clear guidelines for adaptable hyper-parameter settings. Additionally, we demonstrated that selecting the matrices to adapt and determining the corresponding rank to maximize performance using LoRA modules is not trivial. We believe these aspects of our work suggest a promising area for future research.

%% file: text/X_appendix.tex
\clearpage
\appendix
\input{tables/few_shot_vit_b16}

\input{tables/few_shot_vit_b32}
\input{tables/few_shot_vit_l14}

%% file: tables/few_shot_vit_b16.tex
\begin{table*}[t]
\caption{Detailed results for the 11 datasets with ViT-B/16 as backbone. Top-1 accuracy averaged over 3 random seeds is reported. Highest value is highlighted in \textbf{bold}, and the second highest is \underline{underlined}.}
\label{tab:vitb16}
\centering
\resizebox{0.9\textwidth}{!}{
\begin{tabular}{llcccccccccccc}
\toprule
Shots & Method & ImageNet & SUN & Aircraft & EuroSAT & Cars & Food & Pets &  Flowers & Caltech & DTD & UCF & Average
\\ \midrule 
\multirow{1}{*}{0} & \textbf{CLIP} {\tiny \textbf{(ICML '21)}} & 66.7 & 62.6 & 24.7 & 47.5 & 65.3 & 86.1 & 89.1 & 71.4 & 92.9 & 43.6 & 66.7 & 65.1  \\

\midrule
\midrule
\multirow{11}{*}{1} 
& CoOp (4) {\tiny \textbf{(IJCV '22)}} & 68.0 & 67.3 & 26.2 & 50.9 & 67.1 & 82.6 & 90.3 & 72.7 & 93.2 & 50.1 & 70.7 & 67.2 \\
& CoOp (16) {\tiny \textbf{(IJCV '22)}} & 65.7 & 67.0 & 20.8 & 56.4 & 67.5 & 84.3 & 90.2 & 78.3 & 92.5 & 50.1 & 71.2 & 67.6 \\ 
& CoCoOp {\tiny \textbf{(CVPR '22)}} & 69.4 & 68.7 & 28.1 & 55.4 & 67.6 & 84.9 & 91.9 & 73.4 & 94.1 & 52.6 & 70.4 & 68.8 \\
& TIP-Adapter-F {\tiny \textbf{(ECCV '22)}} & 69.4 & 67.2 & 28.8 & \underline{67.8} & 67.1 & 85.8 & 90.6 & \textbf{83.8} & 94.0 & 51.6 & 73.4 & \underline{70.9} \\
& CLIP-Adapter {\tiny \textbf{(IJCV '23)}} & 67.9 & 65.4 & 25.2 & 49.3 & 65.7 & 86.1 & 89.0 & 71.3 & 92.0 & 44.2 & 66.9 & 65.7 \\
& PLOT++ {\tiny \textbf{(ICLR '23)}} & 66.5 & 66.8 & 28.6 & 65.4 & \underline{68.8} & \underline{86.2} & 91.9 & 80.5 & \textbf{94.3} & \textbf{54.6} & \underline{74.3} & 70.7 \\
& KgCoOp {\tiny \textbf{(CVPR '23)}} & 68.9 & 68.4 & 26.8 & 61.9 & 66.7 & \textbf{86.4} & \underline{92.1} & 74.7 & \underline{94.2} & 52.7 & 72.8 & 69.6 \\
& TaskRes {\tiny \textbf{(CVPR '23)}} & 69.6 & 68.1 & \textbf{31.3} & 65.4 & \underline{68.8} & 84.6 & 90.2 & 81.7 & 93.6 & 53.8 & 71.7 & 70.8 \\
& MaPLe {\tiny \textbf{(CVPR '23)}} & \underline{69.7} & \underline{69.3} & 28.1 & 29.1 & 67.6 & 85.4 & 91.4 & 74.9 & 93.6 & 50.0 & 71.1 & 66.4 \\

& ProGrad {\tiny \textbf{(ICCV '23)}} & 67.0 & 67.0 & 28.8 & 57.0 & 68.2 & 84.9 & 91.4 & 80.9 & 93.5 & 52.8 & 73.3 & 69.5  \\

\rowcolor{LightGray} & CLIP-LoRA (Ours) & \textbf{70.4} & \textbf{70.4} & \underline{30.2} & \textbf{72.3} & \textbf{70.1} & 84.3 & \textbf{92.3} & \underline{83.2} & 93.7 & \underline{54.3} & \textbf{76.3} & \textbf{72.5} \\
\midrule
\midrule
\multirow{11}{*}{2}
& CoOp (4) {\tiny \textbf{(IJCV '22)}} & 68.7 & 68.0 & 28.1 & 66.2 & 70.5 & 82.6 & 89.9 & 80.9 & 93.0 & 53.7 & 73.5 & 70.5 \\
& CoOp (16) {\tiny \textbf{(IJCV '22)}} & 67.0 & 67.0 & 25.9 & 65.1 & 70.4 & 84.4 & 89.9 & 88.0 & 93.1 & 54.1 & 74.1 & 70.8 \\
& CoCoOp {\tiny \textbf{(CVPR '22)}} & 70.1 & 69.4 & 29.3 & 61.8 & 68.4 & 85.9 &  \underline{91.9} & 77.8 & 94.4 & 52.3 & 73.4 & 70.4 \\
.& TIP-Adapter-F {\tiny \textbf{(ECCV '22)}} & 70.0 & 68.6 & \underline{32.8} & 73.2 & 70.8 & 86.0 & 91.6 & \textbf{90.1} & 93.9 & \underline{57.8} & 76.2 & 73.7 \\
& CLIP-Adapter {\tiny \textbf{(IJCV '23)}} & 68.2 & 67.2 & 27.0 & 51.2 & 66.6 & 86.2 & 89.7 & 71.7 & 93.4 & 45.4 & 68.4 & 66.8 \\
& PLOT++ {\tiny \textbf{(ICLR '23)}} & 68.3 & 68.1 & 31.1 & \underline{76.8} & \textbf{73.2} & 86.3 & \textbf{92.3} & \underline{89.8} & \underline{94.7} & 56.7 & \underline{76.8} & \underline{74.0} \\
& KgCoOp {\tiny \textbf{(CVPR '23)}} & 69.6 & 69.6 & 28.0 & 69.2 & 68.2 & \textbf{86.6} & \textbf{92.3} & 79.8 & 94.5 & 55.3 & 74.6 & 71.6 \\
& TaskRes {\tiny \textbf{(CVPR '23)}} & \underline{70.2} & 70.5 & 32.7 & 70.2 & 72.1 & 85.6 & 90.7 & 84.4 & 94.3 & 55.6 & 75.2 & 72.9 \\
& MaPLe {\tiny \textbf{(CVPR '23)}} & 70.0 & \underline{70.7} & 29.5 & 59.4 & 68.5 & \underline{86.5} & 91.8 & 79.8 & \textbf{94.9} & 50.6 & 74.0 & 70.5  \\

& ProGrad {\tiny \textbf{(ICCV '23)}} & 69.1 & 69.0 & 31.1 & 66.3 & \underline{72.4} & 84.8 & 91.5 & 87.5 & 93.6 & 56.0 & 75.6 & 72.4 \\
\rowcolor{LightGray} & CLIP-LoRA (Ours) & \textbf{70.8} & \textbf{71.3} & \textbf{33.2} & \textbf{82.7} & \textbf{73.2} & 83.2 & 91.3 & \underline{89.8} & 94.6 & \textbf{59.9} & \textbf{80.0} & \textbf{75.5} \\
\midrule
\midrule
\multirow{11}{*}{4}
& CoOp (4)  {\tiny \textbf{(IJCV '22)}} &  69.7 & 70.6 & 29.7 & 65.8 & 73.4 & 83.5 & 92.3 & 86.6 & 94.5 & 58.5 & 78.1 & 73.0 \\
& CoOp (16)  {\tiny \textbf{(IJCV '22)}} & 68.8 & 69.7 & 30.9 & 69.7 & 74.4 & 84.5 & 92.5  & 92.2 & 94.5 & 59.5 & 77.6 & 74.0 \\
& CoCoOp {\tiny \textbf{(CVPR '22)}} & 70.6 & 70.4 & 30.6 & 61.7 & 69.5 & 86.3 & \underline{92.7} & 81.5 & 94.8 & 55.7 & 75.3 & 71.7 \\
& TIP-Adapter-F {\tiny \textbf{(ECCV '22)}} & 70.7 & 70.8 & \underline{35.7} & 76.8 & 74.1 & 86.5 & 91.9 & 92.1 & 94.8 & 59.8 & 78.1 & 75.6 \\
& CLIP-Adapter {\tiny \textbf{(IJCV '23)}} & 68.6 & 68.0 & 27.9 & 51.2 & 67.5 & 86.5 & 90.8 & 73.1 & 94.0 & 46.1 & 70.6 & 67.7 \\
& PLOT++ {\tiny \textbf{(ICLR '23)}} & 70.4 & 71.7 & 35.3 & \underline{83.2} & \underline{76.3} & 86.5 & 92.6 & \underline{92.9} & \underline{95.1} & \underline{62.4} & \underline{79.8} & \underline{76.9} \\
& KgCoOp {\tiny \textbf{(CVPR '23)}} & 69.9 & 71.5 & 32.2 & 71.8 & 69.5 & \textbf{86.9} & 92.6 & 87.0 & 95.0 & 58.7 & 77.6 & 73.9 \\
& TaskRes {\tiny \textbf{(CVPR '23)}} & \underline{71.0} & \underline{72.7} & 33.4 & 74.2 & 76.0 & 86.0 & 91.9 & 85.0 & 95.0 & 60.1 & 76.2 & 74.7 \\
& MaPLe {\tiny \textbf{(CVPR '23)}} & 70.6 & 71.4 & 30.1 & 69.9 & 70.1 & \underline{86.7} & \textbf{93.3} & 84.9 & 95.0 & 59.0 & 77.1 & 73.5  \\
& ProGrad {\tiny \textbf{(ICCV '23)}} & 70.2 & 71.7 & 34.1 & 69.6 & 75.0 & 85.4 & 92.1 & 91.1 & 94.4 & 59.7 & 77.9 & 74.7 \\

\rowcolor{LightGray} & CLIP-LoRA (Ours) & \textbf{71.4} & \textbf{72.8} & \textbf{37.9} & \textbf{84.9} & \textbf{77.4} & 82.7 & 91.0 & \textbf{93.7} & \textbf{95.2} & \textbf{63.8} & \textbf{81.1} & \textbf{77.4} \\
\midrule
\midrule
\multirow{11}{*}{8}
& CoOp (4) {\tiny \textbf{(IJCV '22)}} & 70.8 & 72.4 & 37.0 & 74.7 & 76.8 & 83.3 & 92.1 & 95.0 & 94.7 & 63.7 & 79.8 & 76.4 \\
& CoOp (16) {\tiny \textbf{(IJCV '22)}} & 70.6 & 71.9 & 38.5 & 77.1 & 79.0 & 82.7 & 91.3 & 94.9 & 94.5 & 64.8 & 80.0 & 76.8 \\
& CoCoOp {\tiny \textbf{(CVPR '22)}} & 70.8 & 71.5 & 32.4 & 69.1 & 70.4 & \underline{87.0} & \textbf{93.3} & 86.3 & 94.9 & 60.1 & 75.9 & 73.8 \\
& TIP-Adapter-F {\tiny \textbf{(ECCV '22)}} & \underline{71.7} & 73.5 & 39.5 & 81.3 & 78.3 & 86.9 & 91.8 & 94.3 & 95.2 & \underline{66.7} & 82.0 & 78.3 \\
& CLIP-Adapter {\tiny \textbf{(IJCV '23)}} & 69.1 & 71.7 & 30.5 & 61.6 & 70.7 & 86.9 & 91.9 & 83.3 & 94.5 & 50.5 & 76.2 & 71.5 \\
& PLOT++ {\tiny \textbf{(ICLR '23)}} & 71.3 & 73.9 & \underline{41.4} & \underline{88.4} & \underline{81.3} & 86.6 & 93.0 & 95.4 & \underline{95.5} & 66.5 & \underline{82.8} & \underline{79.6} \\
& KgCoOp {\tiny \textbf{(CVPR '23)}} & 70.2 & 72.6 & 34.8 & 73.9 & 72.8 & \underline{87.0} & 93.0 & 91.5 & 95.1 & 65.6 & 80.0 & 76.0 \\
& TaskRes {\tiny \textbf{(CVPR '23)}} & \textbf{72.3} & \underline{74.6} & 40.3 & 77.5 & 79.6 & 86.4 & 92.0 & \underline{96.0} & 95.3 & \underline{66.7} & 81.6 & 78.4 \\
& MaPLe {\tiny \textbf{(CVPR '23)}} & 71.3 & 73.2 & 33.8 & 82.8 & 71.3 & \textbf{87.2} & \underline{93.1} & 90.5 & 95.1 & 63.0 & 79.5 & 76.4 \\

& ProGrad {\tiny \textbf{(ICCV '23)}} & 71.3 & 73.0 & 37.7 & 77.8 & 78.7 & 86.1 & 92.2 & 95.0 & 94.8 & 63.9 & 80.5 & 77.4 \\
\rowcolor{LightGray}& CLIP-LoRA (Ours) &  \textbf{72.3} & \textbf{74.7} & \textbf{45.7} & \textbf{89.7} & \textbf{82.1} & 83.1 & 91.7 & \textbf{96.3} & \textbf{95.6} & \textbf{67.5} & \textbf{84.1} & \textbf{80.3}  \\
\midrule
\midrule
\multirow{11}{*}{16}
& CoOp (4)  {\tiny \textbf{(IJCV '22)}} &  71.5 & 74.6 & 40.1 & 83.5 & 79.1 & 85.1 & 92.4 & 96.4 & 95.5 & 69.2 & 81.9 & 79.0 \\
& CoOp (16) {\tiny \textbf{(IJCV '22)}} & 71.9 & 74.9 & 43.2 & 85.0 & 82.9 & 84.2 & 92.0 & 96.8 & 95.8 & 69.7 & 83.1 & 80.0 \\
& CoCoOp {\tiny \textbf{(CVPR '22)}} & 71.1 & 72.6 & 33.3 & 73.6 & 72.3 & \textbf{87.4} & \underline{93.4} & 89.1 & 95.1 & 63.7 & 77.2 & 75.4 \\
& TIP-Adapter-F {\tiny \textbf{(ECCV '22)}} & \underline{73.4} &  \underline{76.0} & 44.6 & 85.9 & 82.3 & 86.8 & 92.6 & 96.2 & 95.7 & 70.8 & 83.9 & 80.7 \\
& CLIP-Adapter {\tiny \textbf{(IJCV '23)}} & 69.8 & 74.2 & 34.2 & 71.4 & 74.0 & 87.1 & 92.3 & 92.9 & 94.9 & 59.4 & 80.2 & 75.5  \\
& PLOT++ {\tiny \textbf{(ICLR '23)}} & 72.6 & \underline{76.0} & \underline{46.7} & \underline{92.0} & \underline{84.6} & 87.1 & \textbf{93.6} & \underline{97.6} & \underline{96.0} & 71.4 & \underline{85.3} & \underline{82.1} \\
& KgCoOp {\tiny \textbf{(CVPR '23)}} & 70.4 & 73.3 & 36.5 & 76.2 & 74.8 & \underline{87.2} & 93.2 & 93.4 & 95.2 & 68.7 & 81.7 & 77.3 \\
& TaskRes {\tiny \textbf{(CVPR '23)}} & 73.0 & \textbf{76.1} & 44.9 & 82.7 & 83.5 & 86.9 & 92.4 & 97.5 & 95.8 & \textbf{71.5} & 84.0 & 80.8 \\
& MaPLe {\tiny \textbf{(CVPR '23)}} &  71.9 & 74.5 & 36.8 & 87.5 & 74.3 & \textbf{87.4} & 93.2 & 94.2 & 95.4 & 68.4 & 81.4 & 78.6 \\
& ProGrad {\tiny \textbf{(ICCV '23)}} &  72.1 & 75.1 & 43.0 & 83.6 & 82.9 & 85.8 & 92.8 & 96.6 & 95.9 & 68.8 & 82.7 & 79.9 \\

\rowcolor{LightGray} & CLIP-LoRA (Ours) & \textbf{73.6} & \textbf{76.1} & \textbf{54.7} & \textbf{92.1}  & \textbf{86.3} & 84.2 & 92.4 & \textbf{98.0} & \textbf{96.4} & \textbf{72.0} & \textbf{86.7} & \textbf{83.0} \\
\bottomrule

\end{tabular}}

\end{table*}

%% file: tables/few_shot_vit_b32.tex
\begin{table*}[t!]
\caption{Detailed results for the 11 datasets with ViT-B/32 as backbone. Top-1 accuracy averaged over 3 random seeds is reported. Highest value is highlighted in \textbf{bold}, and the second highest is \underline{underlined}.}
\label{tab:vitb32}
\centering
\resizebox{0.9\textwidth}{!}{
\begin{tabular}{llcccccccccccc}
\toprule
Shots & Method & ImageNet & SUN & Aircraft & EuroSAT & Cars & Food & Pets &  Flowers & Caltech & DTD & UCF & Average
\\ \midrule 
\multirow{1}{*}{0} & \textbf{CLIP} {\tiny \textbf{(ICML '21)}} & 61.9 & 62.0 & 19.3 & 45.1 & 60.4 & 80.5 & 87.5 & 67.0 & 91.1 & 42.6 & 62.2 & 61.8  \\

\midrule
\midrule
\multirow{11}{*}{1} 
& CoOp (4) {\tiny \textbf{(IJCV '22)}} &  62.7 & 64.8 & 18.7 & 49.3 & 60.5 & 74.4 & 87.7 & 68.0 & 91.5 & 47.1 & 66.5 & 62.8 \\
& CoOp (16) {\tiny \textbf{(IJCV '22)}} & 60.8 & 63.3 & 15.7 & 52.9 & 59.6 & 75.9 & 87.6 & 71.5 & 91.7 & 47.2 & 66.3 & 63.0 \\ 
& CoCoOp {\tiny \textbf{(CVPR '22)}} &  64.3 & 65.6 & 15.5 & 46.7 & 60.4 & 79.5 & 88.2 & 68.6 & 92.2 & 45.7 & 67.3 & 63.1 \\
& TIP-Adapter-F {\tiny \textbf{(ECCV '22)}} & 64.5 & 65.4 & \underline{22.2} & 59.7 & 61.1 & 80.4 & 87.5 & \textbf{81.1} & \underline{92.4} & \underline{50.9} & 66.5 & \underline{66.5} \\
& CLIP-Adapter {\tiny \textbf{(IJCV '23)}} &  63.2 & 63.2 & 19.8 & 46.9 & 60.7 & \underline{80.6} & 87.6 & 66.7 & 91.6 & 44.1 & 62.9 & 62.5 \\
& PLOT++ {\tiny \textbf{(ICLR '23)}} &  60.9 & 64.5 & 22.0 & \textbf{67.3} & \underline{61.4} & 79.2 & \textbf{89.4} & 73.7 & \textbf{93.0} & 50.6 & \underline{69.7} & \underline{66.5} \\
& KgCoOp {\tiny \textbf{(CVPR '23)}} & 63.9 & 65.7 & 20.9 & 55.4 & 61.2 & \textbf{80.7} & 88.6 & 68.1 & 92.2 & 50.2 & 66.8 & 64.9 \\
& TaskRes {\tiny \textbf{(CVPR '23)}} &  \underline{64.6} & 62.0 & 20.9 & 60.1 & 59.6 & 74.6 & 84.3 & 75.2 & 88.8 & 50.1 & 64.3 & 64.0\\
& MaPLe {\tiny \textbf{(CVPR '23)}} & \underline{64.6} & \underline{66.9} & 13.7 & 32.3 & 60.1 & 79.7 & 87.6 & 68.7 & 92.0 & 44.5 & 66.4 & 61.5\\
& ProGrad {\tiny \textbf{(ICCV '23)}} &   62.0 & 64.9 & 21.2 & 52.8 & 60.6 & 78.1 & 87.9 & 74.9 & 91.6 & \textbf{51.0} & 66.5 & 64.7 \\

\rowcolor{LightGray} & CLIP-LoRA (Ours) & \textbf{65.2} & \textbf{67.6} & \textbf{22.9} & \underline{67.1} & \textbf{63.5} &  78.4 & \underline{88.7} & \underline{77.4} & \textbf{93.0} & 49.8 & \textbf{72.3} & \textbf{67.8}   \\
\midrule
\midrule
\multirow{11}{*}{2}
& CoOp (4) {\tiny \textbf{(IJCV '22)}} & 63.3 & 65.6 & 20.8 & 56.5 & 62.1 & 74.2 & 86.6 & 77.1 & 91.5 & 49.4 & 71.5 & 65.3 \\
& CoOp (16) {\tiny \textbf{(IJCV '22)}} &  61.3 & 63.8 & 18.6 & 62.4 & 62.3 & 73.3 & 85.8 & 82.0 & 91.2 & 49.6 & 70.4 & 65.5 \\
& CoCoOp {\tiny \textbf{(CVPR '22)}} & 64.8 & 66.8 & 15.4 & 51.7 & 61.1 & 80.8 & 88.7 & 70.0 & 92.7 & 50.3 & 70.0 & 64.8 \\
& TIP-Adapter-F {\tiny \textbf{(ECCV '22)}} & 64.9 & 66.8 & \textbf{24.6} & 67.4 & 63.7 & 80.8 & 88.2 & \textbf{85.5} & 92.8 & \underline{54.9} & 71.8 & \underline{69.2}  \\
& CLIP-Adapter {\tiny \textbf{(IJCV '23)}} &  63.4 & 64.8 & 20.9 & 50.1 & 61.5 & \underline{81.0} & 87.9 & 67.1 & 91.4 & 44.9 & 65.9 & 63.5\\
& PLOT++ {\tiny \textbf{(ICLR '23)}} & 62.6 & 65.5 & 23.1 & \underline{73.0} & \underline{64.7} & 77.0 & 87.4 & 84.3 & 92.9 & 52.8 & \underline{72.6} & 68.7\\
& KgCoOp {\tiny \textbf{(CVPR '23)}} & 64.2 & 67.5 & 21.7 & 58.5 & 62.4 & \textbf{81.1} & \underline{88.8} & 73.7 & 92.8 & 53.1 & 69.2 & 66.6 \\
& TaskRes {\tiny \textbf{(CVPR '23)}} & \underline{65.2} & 64.4 & 22.4 & 64.8 & 62.8 & 75.9 & 85.2 & 78.0 & 89.9 & 53.6 & 67.8 & 66.4 \\
& MaPLe {\tiny \textbf{(CVPR '23)}} & \underline{65.2} & \underline{67.6} & 20.5 & 46.9 & 61.9 & 80.7 & \textbf{89.3} & 71.6 & \underline{93.0} & 49.7 & 71.0 & 65.2  \\
& ProGrad {\tiny \textbf{(ICCV '23)}} & 63.7 & 67.5 & 22.3 & 56.7 & 64.0 & 78.3 & 87.0 & 82.2 & 92.4 & 52.1 & 71.4 & 67.1 \\
\rowcolor{LightGray} & CLIP-LoRA (Ours) & \textbf{65.7} & \textbf{68.6} & \underline{24.1} & \textbf{80.8} & \textbf{64.8} & 76.3 & 86.6 & \underline{84.7} &  \textbf{93.7} & \textbf{57.4} & \textbf{75.4} & \textbf{70.7}\\
\midrule
\midrule
\multirow{11}{*}{4}
& CoOp (4)  {\tiny \textbf{(IJCV '22)}} & 64.8 & 68.2 & 22.1 & 62.1 & 65.2 & 76.5 & \underline{90.1} & 84.1 & 93.1 & 55.3 & 73.5 & 68.6 \\
& CoOp (16)  {\tiny \textbf{(IJCV '22)}} &  63.2 & 67.1 & 24.0 & 68.7 & 66.2 & 75.6 & 88.8 & 87.9 & 93.0 & 55.3 & 75.0 & 69.5 \\
& CoCoOp {\tiny \textbf{(CVPR '22)}} &  65.2 & 67.8 & 17.3 & 58.5 & 62.0 & 81.1 & 89.8 & 74.6 & 93.2 & 52.3 & 71.6 & 66.7 \\
& TIP-Adapter-F {\tiny \textbf{(ECCV '22)}} & \underline{65.8} & 68.3 &  \textbf{28.8} & 71.5 & 67.6 & 80.9 & 88.6 & 88.9 & \textbf{94.6} & \underline{58.0} & 75.1 & 71.6 \\
& CLIP-Adapter {\tiny \textbf{(IJCV '23)}} &  63.7 & 65.6 & 21.3 & 49.9 & 62.2 & \underline{81.3} & 88.4 & 68.3 & 92.0 & 47.2 & 67.3 & 64.3 \\
& PLOT++ {\tiny \textbf{(ICLR '23)}} & 64.6 & 69.2 & 26.2 & \underline{81.6} & \textbf{68.5} & 77.8 & 89.1 & \textbf{90.2} & 93.9 & 57.2 & \underline{75.6} & \underline{72.2} \\
& KgCoOp {\tiny \textbf{(CVPR '23)}} &  64.7 & 69.2 & 22.6 & 64.9 & 63.2 & 81.2 & 89.5 & 76.8 & 93.8 & 55.1 & 71.6 & 68.4 \\
& TaskRes {\tiny \textbf{(CVPR '23)}} &  66.1 & 66.7 & 23.1 & 70.7 & 66.7 & 76.7 & 86.7 & 79.0 & 90.6 & 57.0 & 68.2 & 68.3 \\
& MaPLe {\tiny \textbf{(CVPR '23)}} &  65.6 & 69.4 & 23.4 & 64.7 & 62.2 & \textbf{81.4} & \textbf{90.5} & 78.1 & 94.0 & 55.0 & 70.9 & 68.7 \\
& ProGrad {\tiny \textbf{(ICCV '23)}} &  65.2 & \underline{69.6} & 24.8 & 63.7 & 66.4 & 79.2 & 89.4 & 87.5 & 93.2 & 55.9 & 73.4 & 69.8 \\

\rowcolor{LightGray} & CLIP-LoRA (Ours) & \textbf{66.5} & \textbf{70.3} & \underline{27.7} & \textbf{85.6} & \underline{68.3} & 75.6 & 86.3 & \underline{90.1} & \underline{94.3} & \textbf{60.3} & \textbf{76.5} & \textbf{72.9} \\
\midrule
\midrule
\multirow{11}{*}{8}
& CoOp (4) {\tiny \textbf{(IJCV '22)}} & 65.8 & 70.0 & 28.0 & 72.8 & 69.0 & 77.2 & 89.5 & 90.2 & 93.8 & 60.8 & 77.1 & 72.2\\
& CoOp (16) {\tiny \textbf{(IJCV '22)}} & 65.5 & 69.2 & 29.1 & 76.4 & 71.3 & 76.3 & 87.4 & 92.7 & 93.8 & 61.7 & 76.5 & 72.7 \\
& CoCoOp {\tiny \textbf{(CVPR '22)}} & 65.8 & 68.9 & 20.3 & 58.1 & 63.4 & 81.6 & 90.1 & 77.3 & 93.8 & 57.4 & 72.4 & 68.1\\
& TIP-Adapter-F {\tiny \textbf{(ECCV '22)}} & 66.8 & 71.2 &  \underline{32.1} & 75.0 & 72.6 & 81.3 & 89.8 & 90.4 & \underline{94.5} & 63.6 & 78.0 & 74.1 \\
& CLIP-Adapter {\tiny \textbf{(IJCV '23)}} & 64.2 & 69.3 & 23.5 & 55.2 & 65.4 & 81.5 & 89.3 & 78.0 & 93.9 & 50.8 & 73.0 & 67.6 \\
& PLOT++ {\tiny \textbf{(ICLR '23)}} & 66.2 & 71.0 & 31.7 & \underline{87.1} & 73.5 & 78.2 & 88.4 & \textbf{93.8} & 94.4 & 62.9 & \underline{79.1} & \underline{75.1} \\
& KgCoOp {\tiny \textbf{(CVPR '23)}} & 65.1 & 69.5 & 24.7 & 66.2 & 65.0 & \underline{81.7} & \underline{90.3} & 83.1 & \underline{94.5} & 61.1 & 74.7 & 70.5 \\
& TaskRes {\tiny \textbf{(CVPR '23)}} &  \textbf{67.4} & \underline{71.9} & 31.9 & 74.9 & \underline{73.8} & 80.6 & 89.1 & \underline{93.5} & \textbf{94.8} & \textbf{64.5} & 78.4 & 74.6 \\
& MaPLe {\tiny \textbf{(CVPR '23)}} & 66.3 & 70.3 & 25.4 & 79.0 & 63.7 & \textbf{81.9} & \textbf{90.9} & 81.1 & 94.4 & 59.8 & 75.0 & 71.6\\
& ProGrad {\tiny \textbf{(ICCV '23)}} & 66.1 & 71.1 & 29.0 & 73.5 & 71.8 & 80.0 & 89.1 & 92.1 & 94.2 & 62.3 & 75.7 & 73.2\\
\rowcolor{LightGray}& CLIP-LoRA (Ours) & \underline{67.2} & \textbf{72.1} & \textbf{36.1} & \textbf{88.8} & \textbf{74.4} &  76.7 & 87.7 & 92.4 & \textbf{94.8} & \underline{63.7} & \textbf{80.1} & \textbf{75.8} \\
\midrule
\midrule
\multirow{11}{*}{16}
& CoOp (4)  {\tiny \textbf{(IJCV '22)}} &  66.7 & 72.1 & 31.1 & 80.6 & 71.7 & 79.8 & 89.4 & 93.1 & 95.0 & 64.3 & 78.2 & 74.7\\
& CoOp (16) {\tiny \textbf{(IJCV '22)}} & 66.8 & 72.2 & 32.9 & 83.3 & 76.0 & 78.6 & 88.7 & 95.4 & 94.9 & 65.3 & 78.6 & 75.7 \\
& CoCoOp {\tiny \textbf{(CVPR '22)}} & 66.0 & 69.8 & 22.6 & 70.4 & 64.6 & \underline{81.9} & \underline{91.0} & 82.5 & 94.3 & 59.7 & 75.3 & 70.7\\
& TIP-Adapter-F {\tiny \textbf{(ECCV '22)}} & \textbf{68.4} & \textbf{74.1} & 34.8 &  83.4 & 77.0 & 81.7 & 90.4 & 94.3 & 95.1 & 68.0 & 80.5 & 77.1  \\
& CLIP-Adapter {\tiny \textbf{(IJCV '23)}} & 64.9 & 71.8 & 26.7 & 64.7 & 68.9 & \underline{81.9} & 90.1 & 88.7 & 94.8 & 58.1 & 76.5 & 71.6 \\
& PLOT++ {\tiny \textbf{(ICLR '23)}} & 67.4 & 73.4 & 36.3 & \underline{91.1} & 77.4 & 79.7 & 89.1 & \textbf{96.3} & 94.9 & 67.0 & \underline{81.5} & \underline{77.6}\\
& KgCoOp {\tiny \textbf{(CVPR '23)}} &  65.4 & 71.0 & 23.7 & 70.1 & 67.3 & 81.7 & 90.8 & 86.1 & 94.4 & 65.1 & 77.5 & 72.1 \\
& TaskRes {\tiny \textbf{(CVPR '23)}} & \underline{68.2} & 73.6 & \underline{37.0} & 77.7 & \underline{78.0} & 81.4 & 89.4 & 95.5 & \textbf{95.7} & \textbf{68.3} & 80.6 & 76.9\\
& MaPLe {\tiny \textbf{(CVPR '23)}} &  66.7 & 72.0 & 28.0 & 83.3 & 66.9 & \textbf{82.1} & \textbf{91.7} & 89.0 & 95.1 & 63.4 & 77.3 & 74.1\\
& ProGrad {\tiny \textbf{(ICCV '23)}} &  66.9 & 73.2 & 33.3 & 81.0 & 76.1 & 80.1 & 89.3 & 95.1 & 95.0 & 65.8 & 79.6 & 75.9 \\
\rowcolor{LightGray} & CLIP-LoRA (Ours) & \textbf{68.4} & \underline{74.0} &  \textbf{44.9} & \textbf{91.8} & \textbf{79.7} & 78.2 & 88.8 & \underline{96.2} &  \underline{95.2} & \underline{68.2} &  \textbf{82.8} & \textbf{78.9} \\
\bottomrule

\end{tabular}}

\end{table*}

%% file: tables/few_shot_vit_l14.tex
\begin{table*}[t!]
\caption{Detailed results for the 11 datasets with ViT-L/14 as backbone. Top-1 accuracy averaged over 3 random seeds is reported. Highest value is highlighted in \textbf{bold}, and the second highest is \underline{underlined}.}
\label{tab:vitl14}
\centering
\resizebox{0.9\textwidth}{!}{
\begin{tabular}{llcccccccccccc}
\toprule
Shots & Method & ImageNet & SUN & Aircraft & EuroSAT & Cars & Food & Pets &  Flowers & Caltech & DTD & UCF & Average
\\ \midrule 
\multirow{1}{*}{0} & \textbf{CLIP} {\tiny \textbf{(ICML '21)}} & 72.9 & 67.6 & 32.6 & 58.0 & 76.8 & 91.0 & 93.6 & 79.4 & 94.9 & 53.6 & 74.2 & 72.2  \\

\midrule
\midrule
\multirow{11}{*}{1} 
& CoOp (4) {\tiny \textbf{(IJCV '22)}} & 73.9 & 70.9 & 35.8 & 64.3 & 78.8 & 89.3 & 93.0 & 86.5 & 93.8 & 58.0 & 77.4 & 74.7 \\
& CoOp (16) {\tiny \textbf{(IJCV '22)}}  & 71.5 & 68.9 & 36.9 & 68.3 & 78.6 & 89.0 & 94.0 & 87.2 & 94.9 & 58.5 & 78.5 & 75.1 \\
& CoCoOp {\tiny \textbf{(CVPR '22)}}  & 76.0 & 72.1 & 36.5 & 65.0 & 78.8 & \underline{90.8} & \textbf{94.7} & 83.0 & \textbf{95.7} & 57.5 & 78.5 & 75.3 \\
& TIP-Adapter-F {\tiny \textbf{(ECCV '22)}} & 76.4 & 71.0 & 38.5 & 67.8 & 79.2 & \textbf{91.0} &  93.2 & \underline{90.9} &  95.3 & 59.3 & 77.9 & 76.4\\
& CLIP-Adapter {\tiny \textbf{(IJCV '23)}}  & 74.6 & 69.3 & 32.9 & 60.1 & 77.2 & \textbf{91.0} & 93.6 & 79.5 & 94.9 & 53.7 & 74.7 & 72.9 \\
& PLOT++ {\tiny \textbf{(ICLR '23)}} & 73.7 & 71.1 & 35.2 & \underline{72.4} & 78.9 & 89.4 & 93.6 & 85.0 & \underline{95.6} & 58.9 & \underline{80.7} & 75.9 \\
& KgCoOp {\tiny \textbf{(CVPR '23)}}  & 75.5 & 72.2 & 35.9 & 69.5 & 77.9 & \textbf{91.0} & 94.1 & 82.2 & 95.5 & \textbf{61.2} & 77.7 & 75.7 \\
& TaskRes {\tiny \textbf{(CVPR '23)}} & 76.2 & 71.4 & \underline{39.7} & 70.6 & \underline{79.8} & 89.9 & 93.5 & 87.6 & 94.6 & \underline{59.8} & 77.4 & 76.4 \\
& MaPLe {\tiny \textbf{(CVPR '23)}}  & \underline{76.5} & \underline{73.3} & 37.4 & 61.2 & 79.1 & 90.2 & 94.4 & 83.6 & 95.2 & 58.4 & 78.5 & 75.3\\
& ProGrad {\tiny \textbf{(ICCV '23)}}  & 73.6 & 71.1 & 38.3 & 70.8 & \underline{79.8} & 90.5 & \underline{94.5} & 88.8 & 95.5 & 58.5 & 80.1 & \underline{76.5} \\

\rowcolor{LightGray} & CLIP-LoRA (Ours) & \textbf{76.7} & \textbf{74.3} & \textbf{41.2} & \textbf{73.7} & \textbf{80.7} & 90.5 & 94.2 &  \textbf{91.2} & \textbf{95.7} & \textbf{61.2} & \textbf{82.4} & \textbf{78.3} \\
\midrule
\midrule
\multirow{11}{*}{2}
& CoOp (4) {\tiny \textbf{(IJCV '22)}} & 74.6 & 70.7 & 38.4 & 71.4 & 81.0 & 89.5 & 93.7 & 92.7 & 95.6 & 60.8 & 81.1 & 77.2 \\
& CoOp (16) {\tiny \textbf{(IJCV '22)}}  & 73.1 & 69.3 & 38.8 & 72.7 & 81.2 & 89.4 & 93.5 & 93.5 & 95.0 & 59.2 & 80.0 & 76.9 \\
& CoCoOp {\tiny \textbf{(CVPR '22)}}  & 76.3 & 73.0 & 38.9 & 72.5 & 79.7 & \underline{91.2} & \textbf{94.4} & 87.1 & 96.2 & 60.7 & 80.8 & 77.3 \\
& TIP-Adapter-F {\tiny \textbf{(ECCV '22)}} & 76.6 &  72.6 & \underline{42.2} & 76.7 & 80.4 & 91.1 & 93.8 & \underline{93.8} & 95.5 & 60.3 & 80.4 & \underline{78.5} \\
& CLIP-Adapter {\tiny \textbf{(IJCV '23)}} & 74.9 & 71.0 & 34.0 & 61.9 & 78.1 & \underline{91.2} & 93.6 & 79.8 & 95.2 & 55.6 & 76.1 & 73.8 \\
& PLOT++ {\tiny \textbf{(ICLR '23)}} & 75.0 & 71.4 & 40.6 & \underline{76.8} & 80.7 & 89.6 & 93.5 & 93.6 & \underline{96.7} & \underline{65.0} & 80.3 &  \underline{78.5} \\
& KgCoOp {\tiny \textbf{(CVPR '23)}}  & 76.0 & 73.9 & 38.1 & 76.5 & 79.3 & \textbf{91.4} & \underline{94.3} & 83.6 & 96.1 & 63.3 & 78.9 & 77.4 \\
& TaskRes {\tiny \textbf{(CVPR '23)}} & 76.6 & 73.1 & 42.0 & 74.4 & \underline{81.7} & 90.6 & 93.8 & 89.8 & 95.7 & 62.5 & 79.2 & 78.1\\
& MaPLe {\tiny \textbf{(CVPR '23)}} & \underline{76.9} & \underline{74.9} & 38.3 & 70.9 & 79.2 & \underline{91.2} & 94.2 & 87.7 & 96.4 & 61.3 & \underline{81.3} & 77.5 \\
& ProGrad {\tiny \textbf{(ICCV '23)}} & 75.1 & 72.6 & 41.1 & 73.7 & \textbf{82.3} & 90.6 & 94.1 & \underline{93.8} & 96.1 & 61.5 & 81.1 & 78.4 \\
\rowcolor{LightGray} & CLIP-LoRA (Ours) & \textbf{77.3} &  \textbf{75.5} & \textbf{42.4} & \textbf{85.9} & \textbf{82.3} & 89.8 & 93.2 & \textbf{94.7} &  \textbf{96.8} & \textbf{67.3} & \textbf{84.2} & \textbf{80.9} \\
\midrule
\midrule
\multirow{11}{*}{4}
& CoOp (4)  {\tiny \textbf{(IJCV '22)}} &   76.0 & 73.7 & 41.8 & 77.9 & 82.6 & 88.8 & 94.7 & 94.9 & 96.1 & 64.3 & 83.6 & 79.5\\
& CoOp (16)  {\tiny \textbf{(IJCV '22)}} &  74.9 & 73.1 & 43.6 & 75.9 & 83.3 & 88.7 & 94.6 & \underline{95.9} & 96.5 & 63.9 & 82.8 & 79.4 \\
& CoCoOp {\tiny \textbf{(CVPR '22)}} &  77.0 & 74.7 & 41.0 & 74.7 & 79.7 & \underline{91.3} & \underline{94.9} & 89.8 & \underline{97.1} & 64.9 & 82.6 & 78.9\\
& TIP-Adapter-F {\tiny \textbf{(ECCV '22)}} & 77.1 & 74.1 & \underline{47.4} & \underline{81.4} & 82.3 & 91.2 & 94.0 &  95.5 & 96.5 & 64.4 & \underline{83.9} & \underline{80.7}\\
& CLIP-Adapter {\tiny \textbf{(IJCV '23)}} &  75.2 & 72.1 & 35.8 & 61.3 & 78.8 & 91.2 & 93.7 & 81.7 & 95.6 & 57.9 & 77.9 & 74.7\\
& PLOT++ {\tiny \textbf{(ICLR '23)}} & 76.4 & 75.2 & 43.2 & 81.3 & 82.6 & 87.7 & 94.2 & \underline{95.9} & 96.9 & \underline{66.8} & 83.8 & 80.4 \\
& KgCoOp {\tiny \textbf{(CVPR '23)}} & 76.4 & 75.2 & 40.6 & 79.5 & 80.0 & \textbf{91.5} & 94.4 & 90.2 & 96.9 & 66.3 & 83.4 & 79.5\\
& TaskRes {\tiny \textbf{(CVPR '23)}} & 77.1 & 74.9 & 42.5 & 76.6 & 83.6 & 90.7 & 94.4 & 90.3 & 96.5 & 65.4 & 80.1 & 79.3 \\
& MaPLe {\tiny \textbf{(CVPR '23)}} &  \underline{77.2} & \underline{76.0} & 40.4 & 74.6 & 80.3 & \textbf{91.5} & \textbf{95.0} & 93.2 & 97.0 & 64.5 & 82.8 & 79.3 \\
& ProGrad {\tiny \textbf{(ICCV '23)}} & 76.5 & 75.0 & 44.6 & 79.3 & \underline{83.8} & 90.6 & 94.8 & 95.6 & 96.8 & 66.3 & 83.6 & 80.6\\

\rowcolor{LightGray} & CLIP-LoRA (Ours) & \textbf{77.9} & \textbf{76.7} &  \textbf{48.9} & \textbf{86.4} & \textbf{85.2} & 89.6 & 93.9 & \textbf{97.4} &  \textbf{97.2} & \textbf{70.4} & \textbf{86.4} & \textbf{82.7} \\
\midrule
\midrule
\multirow{11}{*}{8}
& CoOp (4) {\tiny \textbf{(IJCV '22)}} & 77.2 & 75.5 & 48.7 & 81.4 & 85.9 & 89.2 & 94.5 & 97.5 & 96.5 & 68.8 & 86.0 & 81.9\\
& CoOp (16) {\tiny \textbf{(IJCV '22)}} & 76.8 & 75.0 & \underline{51.2} & 82.8 & \underline{86.4} & 88.6 & 94.0 & \textbf{98.0} & 96.7 & 69.4 & 85.1 & 82.2  \\
& CoCoOp {\tiny \textbf{(CVPR '22)}} & 77.4 & 75.6 & 43.3 & 77.0 & 81.4 & \textbf{91.6} & \textbf{95.3} & 93.0 & \underline{97.0} & 67.9 & 84.5 & 80.4 \\
& TIP-Adapter-F {\tiny \textbf{(ECCV '22)}} &  77.8 & 76.7 &  50.4 & 84.9 & 85.9 & \underline{91.4} & 94.1 & 97.3 & 96.9 & \underline{71.2} &  \underline{86.2} & \underline{83.0} \\
& CLIP-Adapter {\tiny \textbf{(IJCV '23)}} &  75.7 & 75.9 & 40.7 & 67.9 & 81.6 & \underline{91.4} & 94.3 & 92.3 & 96.8 & 63.8 & 82.8 & 78.5 \\
& PLOT++ {\tiny \textbf{(ICLR '23)}} & 77.8 & 77.0 & 43.2 & \underline{87.0} & 84.6 & 89.6 & 93.3 & 96.3 & 96.8 & 69.5 & 84.8 & 81.8 \\
& KgCoOp {\tiny \textbf{(CVPR '23)}} &  76.7 & 76.2 & 45.9 & 82.1 & 82.3 & \textbf{91.6} & \underline{95.1} & 95.2 & \textbf{97.3} & 70.8 & 85.7 & 81.7\\
& TaskRes {\tiny \textbf{(CVPR '23)}} &  77.9 & 76.0 & 51.1 & 81.1 & 85.7 & 91.1 & 94.5 & 96.7 & 96.9 & 69.4 & 85.6 & 82.4\\
& MaPLe {\tiny \textbf{(CVPR '23)}} & \underline{78.0} & \underline{77.2} & 42.9 & 80.7 & 81.8 & 90.1 & 95.0 & 95.8 & 96.8 & 69.5 & 85.1 & 81.2 \\
& ProGrad {\tiny \textbf{(ICCV '23)}} &  77.7 & 76.1 & 49.9 & 83.6 & 86.2 & 90.8 & \underline{95.1} & \underline{97.8} & 96.7 & 69.9 & 85.4 & 82.7 \\
\rowcolor{LightGray}& CLIP-LoRA (Ours) &  \textbf{78.5} & \textbf{78.0} & \textbf{57.5} & \textbf{90.0} & \textbf{88.7} & 89.7 &  94.2 & \textbf{98.0} & \underline{97.0} & \textbf{72.2} & \textbf{88.3} & \textbf{84.7} \\
\midrule
\midrule
\multirow{11}{*}{16}
& CoOp (4)  {\tiny \textbf{(IJCV '22)}} & 78.1 & 77.9 & 53.0 & 86.7 & 87.4 & 90.2 & 94.5 & 98.6 & \textbf{97.5} & 73.7 & 86.7 & 84.0\\
& CoOp (16) {\tiny \textbf{(IJCV '22)}} & 78.2 & 77.5 & 55.2 & 88.3 & \underline{89.0} & 89.8 & 94.6 & \textbf{99.1} & 97.2 & 74.4 & 87.3 & 84.6 \\
& CoCoOp {\tiny \textbf{(CVPR '22)}} & 77.8 & 76.7 & 45.2 & 79.8 & 82.7 & \underline{91.9} & \textbf{95.4} & 95.3 & \underline{97.4} & 71.4 & 85.2 & 81.7 \\
& TIP-Adapter-F {\tiny \textbf{(ECCV '22)}} &  \underline{79.3} & \textbf{79.6} & \underline{55.8} & 86.1 & 88.1 & 91.6 & 94.6 & 98.3 & \textbf{97.5} & 74.0 & 87.4 & 84.8 \\
& CLIP-Adapter {\tiny \textbf{(IJCV '23)}} &  76.4 & 78.0 & 46.4 & 75.8 & 83.8 & 91.6 & 94.3 & 97.3 & 97.3 & 71.3 & 86.1 & 81.7\\
& PLOT++ {\tiny \textbf{(ICLR '23)}} & 78.6 & 79.1 & 44.1 & \underline{92.2} & 87.2 & 90.2 & 93.6 & 98.8 & \textbf{97.5} & \underline{75.0} & 87.1 & 83.9 \\
& KgCoOp {\tiny \textbf{(CVPR '23)}} &  76.8 & 76.7 & 47.5 & 83.6 & 83.2 & 91.7 & \underline{95.3} & 96.4 & \underline{97.4} & 73.6 & 86.4 & 82.6 \\
& TaskRes {\tiny \textbf{(CVPR '23)}} &  78.1 & 76.9 & 55.0 & 84.3 & 87.6 & 91.5 & 94.7 & 97.8 & 97.3 & 74.4 & 86.6 & 84.0 \\
& MaPLe {\tiny \textbf{(CVPR '23)}} & 78.4 & 78.8 & 46.3 & 85.4 & 83.6 & \textbf{92.0} & \textbf{95.4} & 97.4 & 97.2 & 72.7 & 86.5 & 83.1 \\
& ProGrad {\tiny \textbf{(ICCV '23)}} & 78.4 & 78.3 & 55.6 & 89.3 & 88.8 & 90.8 & 94.9 & 98.7 & \textbf{97.5} & 73.7 & \underline{87.7} & \underline{84.9} \\

\rowcolor{LightGray} & CLIP-LoRA (Ours) &  \textbf{79.6} & \underline{79.4} & \textbf{66.2} & \textbf{93.1} & \textbf{90.9} & 89.9 & 94.3 & \underline{99.0} & 97.3 & \textbf{76.5} & \textbf{89.9} & \textbf{86.9} \\
\bottomrule

\end{tabular}}

\end{table*}